\documentclass[11pt]{article}

\usepackage[utf8]{inputenc}

\makeatletter
\newlength\aftertitskip     \newlength\beforetitskip
\newlength\interauthorskip  \newlength\aftermaketitskip

\def\maketitle{\par
 \begingroup
   \def\thefootnote{\fnsymbol{footnote}}
   \def\@makefnmark{\hbox to 4pt{$^{\@thefnmark}$\hss}}
   \@maketitle \@thanks
 \endgroup
\setcounter{footnote}{0}
 \let\maketitle\relax \let\@maketitle\relax
 \gdef\@thanks{}\gdef\@author{}\gdef\@title{}\let\thanks\relax}

\def\@startauthor{\noindent \normalsize\bf}
\def\@endauthor{}
\def\@starteditor{\noindent \small {\bf Editor:~}}
\def\@endeditor{\normalsize}
\def\@maketitle{\vbox{\hsize\textwidth
 \linewidth\hsize \vskip \beforetitskip
 {\begin{center} \LARGE\@title \par \end{center}} \vskip \aftertitskip
 {\def\and{\unskip\enspace{\rm and}\enspace}%
  \def\addr{\small\it}%
  \def\email{\hfill\small\tt}%
  \def\name{\normalsize\bf}%
  \def\AND{\@endauthor\rm\hss \vskip \interauthorskip \@startauthor}
  \@startauthor \@author \@endauthor}
}}

\makeatother

\usepackage{amsgen,amsmath,amstext,amsbsy,amsopn,amssymb,amsthm}
\usepackage{array,multirow}
\usepackage{graphicx}
\usepackage[ruled,noend]{algorithm2e}
\usepackage{caption}
\usepackage{subcaption}
\usepackage{enumitem}
\usepackage{dsfont}
\usepackage{nicefrac}
\usepackage{array,multirow}
\usepackage{mathtools}
\usepackage{etoc}
\usepackage[export]{adjustbox}
\usepackage{color}
\usepackage{multicol}
\definecolor{cvprblue}{rgb}{0.21,0.49,0.74}
\usepackage[pagebackref,breaklinks,colorlinks,citecolor=cvprblue]{hyperref}

\DeclareMathOperator*{\argmax}{argmax}

\theoremstyle{definition}

\usepackage{booktabs} 
\usepackage{makecell} 

\numberwithin{equation}{section}

\newcommand{\realNumber}{\mathbb{R}}

\newcommand{\hidden}{h}
\newcommand{\cell}{c}
\newcommand{\weight}{W}
\newcommand{\bias}{b}

\newcommand{\network}{\mathcal{N}}

\newcommand{\class}{c}
\newcommand{\classes}{\mathcal{C}}
\renewcommand{\frame}{t}
\newcommand{\frames}{T}
\newcommand{\image}{I}
\newcommand{\advDomain}{\mathbb{S}}
\newcommand{\linfty}{L_\infty}

\newcommand{\map}{\mathcal{B}}
\newcommand{\decoder}{\mathcal{D}}

\newcommand{\probability}{\Pr}

\title{STR-Cert: Robustness Certification for\\ Deep Text Recognition on \\Deep Learning Pipelines and Vision Transformers}

\author{\name Daqian Shao
\email{daqian.shao@cs.ox.ac.uk}\\
  \addr{University of Oxford}\\
  \name Lukas Fesser \email{lukas\_fesser@fas.harvard.edu}\\
  \addr{Harvard University}\\
  \name Marta Kwiatkowska \email{marta.kwiatkowska@cs.ox.ac.uk}\\
  \addr{University of Oxford}
}

\begin{document}
\maketitle

\begin{abstract}
\noindent Robustness certification, which aims to formally certify the predictions of neural networks against adversarial inputs, has become an important tool for safety-critical applications. Despite considerable progress, existing certification methods are limited to elementary architectures, such as convolutional networks, recurrent networks and recently Transformers, on benchmark datasets such as MNIST. In this paper, we focus on the robustness certification of scene text recognition (STR), which is a complex and extensively deployed image-based sequence prediction problem. We tackle three types of STR model architectures, including the standard STR pipelines and the Vision Transformer. We propose STR-Cert, the first certification method for STR models, by significantly extending the DeepPoly polyhedral verification framework via deriving novel polyhedral bounds and 
algorithms for key STR model components. Finally, we certify and compare STR models on six datasets, demonstrating the efficiency and scalability of robustness certification, particularly for the Vision Transformer.
\end{abstract}


\section{Introduction}
\begin{figure}[t]
\centering
\includegraphics[width=\textwidth]{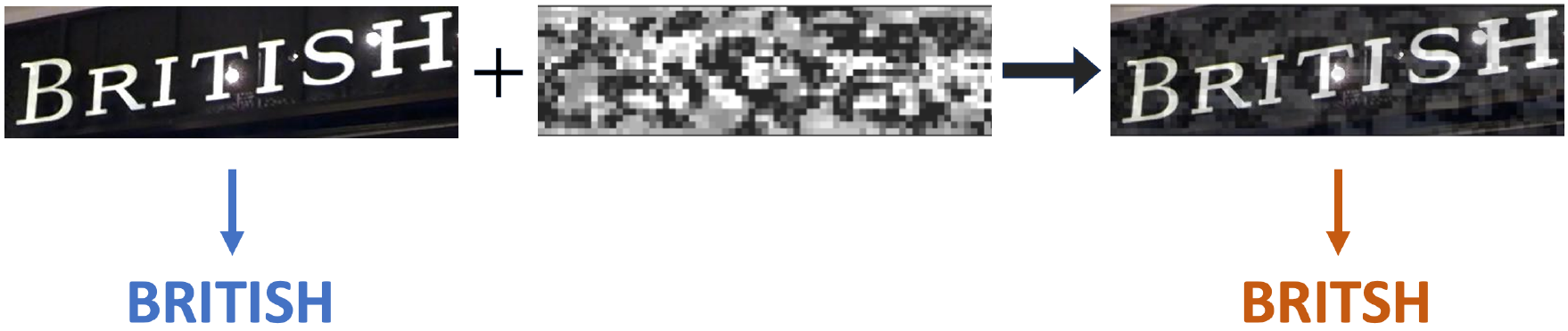}
    \caption{An \textit{adversarial example} from the IC15 dataset, where the predicted text changes under small $\linfty$ perturbations. The original image is also shown to be not robust by STR-Cert.}
    \label{fig:adversarial}
\end{figure}

While deep learning has achieved remarkable performance in a broad range of tasks, such as image classification and natural language processing, 
deep neural networks (DNNs) are known to be vulnerable to adversarial attacks~\cite{Goodfellow2015,Zhang2020,Xu2020AutomaticBeyond}. 
These are inputs that, whilst originally predicted correctly, are misclassified after adding slight and often imperceptible perturbations, illustrated in \ref{fig:adversarial}. The discovery of these vulnerabilities has motivated a multitude of studies on the robustness of neural networks~\cite{Cohen2019CertifiedSmoothing,Zhang2018}. One popular direction 
is neural network \emph{certification} (also called \emph{verification} in the literature), which aims to automatically prove that a network satisfies properties of interest, such as robustness to perturbations~\cite{Tjeng2017EvaluatingProgramming,Zhang2018}, safety guarantees~\cite{akintunde2019verification} and other pre- and post-conditions. Although recent advances 
have made 
certification an important tool for analyzing and reasoning about neural networks, its scalability, precision and support for complex deep learning models and tasks still remain key challenges~\cite{Li2023} in the field.
Most of the existing certification methods focus on simple architectures, such as fully-connected networks (FCNNs) and convolutional networks (CNNs). Some progress has recently been achieved for recurrent networks (RNNs)~\cite{Ryou2020ScalableNetworks,Ko2019}, CNN-RNN models~\cite{Wu2019} and Transformers~\cite{Shi2020RobustnessTransformers,BonaertETHZurich2021FastTransformers,Liao2022AreNetworks}, but typically only for simple tasks and datasets such as MNIST classification (see the taxonomy of Li \textit{et al.}~\cite{Li2023}).
Thus, there is a lack of efficient methodologies and experimental evaluation for robustness certification of complex decision pipelines used in real-world applications on large benchmark datasets, often due to the poor scalability of the underlying techniques.

In this work, we focus on scene text recognition (STR), which is the task of recognizing text from natural images.
STR systems are extensively deployed in businesses, banks, and law enforcement for applications such as document information extraction and number plate recognition, but are prone to adversarial attacks~\cite{Liang2018,Yuan2020AdaptiveRecognition}. Unlike general object recognition, STR is an image-based sequence recognition problem that requires the model to predict a sequence of object labels given an image. To achieve good performance, STR systems are built as sophisticated pipelines, dramatically increasing the complexity and difficulty of the task, precluding direct application of existing methodologies. In this work, we certify three types of STR model architectures including the standard STR architecture pipeline~\cite{Shi2017,Wojna2017Attention-BasedImagery} (see  \ref{sec:STR_arch}) and the Vision Transformer~\cite{Atienza2021VisionRecognition}.
Building on and significantly extending the DeepPoly polyhedral verification~\cite{singh2019abstract,Ryou2020ScalableNetworks} framework, we develop STR-Cert, an efficient 
robustness certification methodology that scales to these complex STR pipelines. Experiments are conducted on six STR datasets to demonstrate the usability of our methods in practice, where we compare and provide insight on the robustness of different STR models.
%
%

\vspace{4pt}
\noindent\textbf{Novel contributions}
Our contributions are as follows.
\begin{itemize}
    \item We propose STR-Cert, an efficient and scalable robustness certification method, and to the best of our knowledge, the first method to certify the robustness of text recognition models and the Vision Transformer.
    \item We derive novel polyhedral bounds and algorithms to certify key components of the STR models such as the CTC decoder, the Softmax function, patch embedding, and the spatial transformation network.
    \item We significantly extend the polyhedral verification framework to implement STR-Cert, which can certify 3 types of STR architectures including the Vision Transformer.
    \item We extensively certify the robustness of the STR models on 6 datasets, providing insights and comparisons between different architectures. 
\end{itemize}
%
%
\vspace{4pt}
\noindent\textbf{Related Works}
\label{sec:related}
Since neural networks can be encoded as sets of constraints, robustness to adversarial perturbations can be solved
exactly and optimally, using mixed integer programming~\cite{Tjeng2017EvaluatingProgramming,BonaertETHZurich2021FastTransformers}, branch and bound~\cite{Wang2021Beta-CROWN:Verification},
or satisfiability modulo theory~\cite{Katz2017Reluplex:Networks}. These methods are referred to as complete verification. However, complete verification is NP-complete even for a simple ReLU network~\cite{Salzer2021}. Another group of methods considers a relaxation to the verification problem: incomplete verification, which guarantees to output ``not verified'' if the input is not safe, but not vice versa.
Many incomplete verification methods can be viewed as applying convex relaxations for non-linear activation functions~\cite{salman2019convex}, including those based on duality~\cite{wong2018provable, dvijotham2018dual}, polyhedron abstraction~\cite{Zhang2018,singh2019abstract,Xu2021FastVerifiers}, zonotope abstraction~\cite{BonaertETHZurich2021FastTransformers}, layer-by-layer reachability analysis ~\cite{wang2018efficient,weng2018towards}, multi-neuron relaxations~\cite{Muller2021} and semi-definite relaxations~\cite{Raghunathan2018}. Robustness verification can also be achieved through an analysis of local Lipschitz constants~\cite{Scaman2018,Jordan2020,Zhang2019RecurJac:Applications}. In this work, we focus on incomplete verification through polyhedral abstraction, which abstracts the input domain and propagates the abstract domain through the neural network via abstract transformations. It is faster than semi-definite and multi-neuron relaxations, and more precise than other weakly relational domains like zonotopes and interval bounds~\cite{Li2023}, offering a good trade-off between scalability and precision. CROWN~\cite{Zhang2018} and DeepPoly~\cite{singh2019abstract} are the first methods to utilize polyhedral abstraction to certify FCNNs, and were further developed into certification methods for CNNs with general activation functions~\cite{Boopathy2019}, RNNs~\cite{Ryou2020ScalableNetworks}, and NLP Transformers~\cite{Shi2020RobustnessTransformers}, including the self-attention mechanism and the Softmax function. In addition to architecture support, recent improvements in scalability and precision include incorporating BaB~\cite{Wang2021Beta-CROWN:Verification,Shi2023}, GPU parallelism~\cite{muller2021scaling}, and optimized polyhedral relaxations~\cite{Xu2021FastVerifiers}.
However, most robustness verification methods are evaluated on MNIST and CIFAR-10, recently scaling up to Tiny ImageNet~\cite{Li2023}, while only a few consider tasks other than image classification. Speech recognition~\cite{Ryou2020ScalableNetworks,Olivier2021} and sentiment analysis~\cite{Shi2020RobustnessTransformers,Ko2019} are studied in the context of RNN and Transformer certification, while existing certification methods have been applied to object detection/segmentation~\cite{Fischer2021} and reinforcement learning~\cite{Bacci2021} problems. To the best of our knowledge, STR-Cert is the first methodology for certifying STR networks --- complex deep learning pipelines with sequence outputs --- based on novel algorithms, which we evaluate on six commonly used STR benchmark datasets.

\section{Background}

Let $\network$ be a neural network for STR tasks with $\frames$ prediction frames and a set $\classes$ of classes that includes all letters and numbers, special symbols, and special tokens such as \textit{blank} and \textit{end-of-sentence} tokens depending on the architecture. Given an image input $\image\in\realNumber^{H \times W \times C}$, where $H, W$ and $C$ are height, weight, and channels of the image, respectively, denote the output of the network by $\network(\image)$, where for each frame $\frame\in[1..\frames]$ and class $\class \in \classes$, $\network_{\frame,\class}(\image)$ is the logits output for predicting class $\class$ in frame $\frame$. The network output $\network(\image)$ is then fed into different decoders $\decoder$ depending on the architecture to retrieve the sequence of characters $\decoder(\network(\image))$ as the text in the image.


\subsection{Robustness Certification for Neural Networks} 

For an input $\image$ and a STR network $\network$ with its decoder $\decoder$, an adversary can perturb the input to a set $\advDomain_\image$ of possible perturbations, denoted as \textit{adversarial region}, before feeding into $\network$. The \textit{adversarial robustness} problem~\cite{singh2019abstract} checks whether, for all possible perturbations $\image^\prime\in\advDomain_\image$, the predicted sequence and that for the original unperturbed input coincide, i.e., $\decoder(\network(\image^\prime))=\decoder(\network(\image))$.  If so, it can be certified that the adversary cannot alter the
classification by picking inputs from $\advDomain_\image$. In this work, we focus on certifying robustness for adversarial regions that can be represented by the Cartesian product of interval constraints, i.e., $\advDomain=\bigtimes_{i=1}^{H\cdot W\cdot C} [l_i,u_i]$ with $l_i,u_i\in\realNumber\cup\{-\infty,\infty\}$, which allows the certification against the widely used $\linfty$-norm attacks up to some \emph{perturbation budget}.
\begin{figure*}[!h]
\centering
\begin{subfigure}{\textwidth}
\centering
\includegraphics[width=0.8\textwidth,fbox]{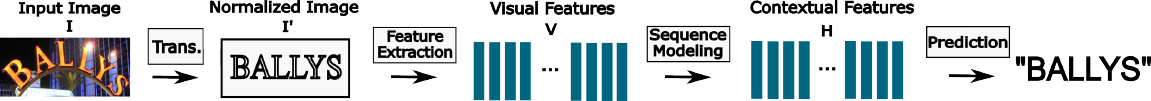}
    \caption{The standard four-stage pipeline for STR.}
    \label{fig:str_pipelines}
    \vspace{5pt}
\end{subfigure}
\begin{subfigure}{\textwidth}
    \centering
\includegraphics[width=0.8\textwidth,fbox]{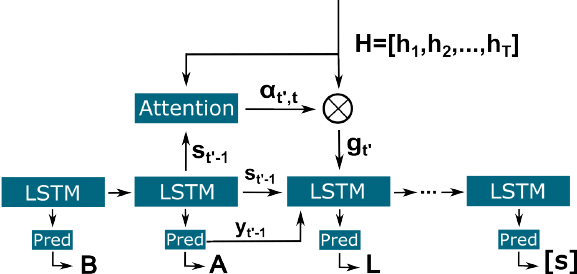}
    \caption{The attention decoder.}
    \label{fig:attention_decoder}
\end{subfigure}
\hfill
\begin{subfigure}{\textwidth}
\centering
\includegraphics[width=0.8\textwidth,fbox]{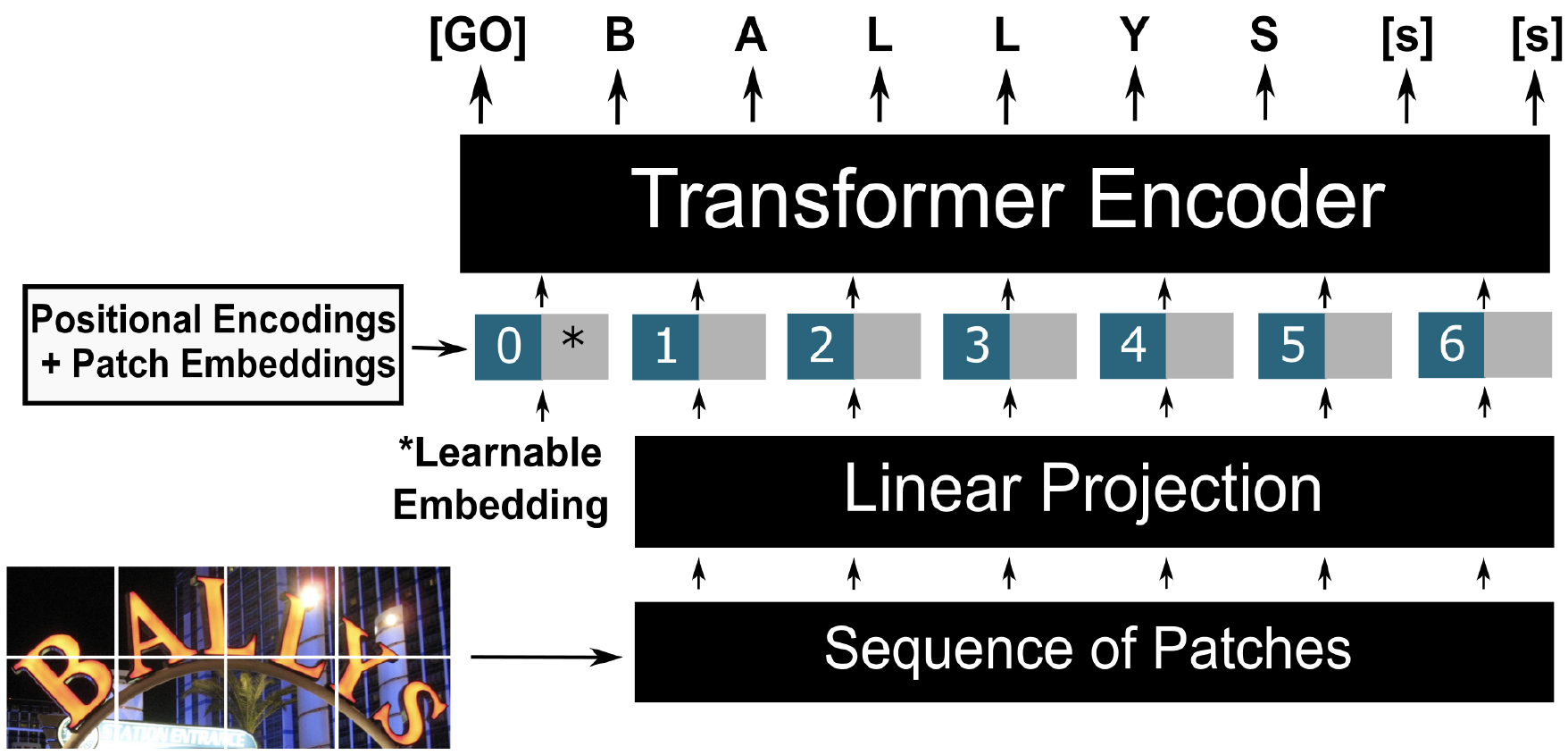}
    \caption{Vision Transformer for scene text recognition (ViTSTR).}
    \label{fig:vitstr}
\end{subfigure}
\caption{Three types of common STR model architectures we consider in this work.}
\end{figure*}
\subsection{Standard STR Architectures}\label{sec:STR_arch}

We consider three STR architectures, of which two are instances of standard STR architectures~\cite{Shi2017,Wojna2017Attention-BasedImagery}, and the third is the popular Vision Transformer~\cite{Atienza2021VisionRecognition}. In this section, we introduce the standard STR architectures, which consist of four stages~\cite{Baek2019} as shown in \ref{fig:str_pipelines}:
\begin{enumerate}
    \item \textbf{Transformation} rectifies and normalizes curved and angled text in the image through a variant of the spatial transformer network (STN~\cite{Jaderberg2015});
    \item  \textbf{Feature extraction} maps the rectified image into a sequence of $\frames$ visual features through a CNN network, denoted by $\mathcal{V} = [v_1, v_2, ..., v_T ]$;
    \item \textbf{Sequence modeling} processes the visual features and produces the contextual features $\mathcal{H} = [h_1, h_2,... , h_T]$ using a long short-term memory (LSTM) network;
    \item \textbf{Prediction} decodes the contextual features to predict the output text. The decoder can be a connectionist temporal classification (CTC) decoder~\cite{Graves2006}, or an attention decoder~\cite{Wojna2017Attention-BasedImagery}, both introduced below.
\end{enumerate}
The first three stages constitute the STR network $\network$ with $\network(\image)=\mathcal{H}$, whereas the last stage is the decoder $\decoder$.

\subsubsection{Transformation Stage}\label{sec:transf}

In STR models, thin plate spline (TPS) transformation~\cite{Shi2016,Liu2016_star}, a variant of STN~\cite{Jaderberg2015}, is typically applied to rectify the input images. Given an image $\image$, TPS adopts a localization CNN network to produce \emph{fiducial points} $\Theta$ that generate the transformation grid. Then, let
\begin{equation}
\mathcal{T} = \left( \Delta_{\Theta^\prime}^{-1} \begin{bmatrix}
\Theta^\top \\ 0 \end{bmatrix} \right)^\top,
\end{equation}
where $\Theta^\prime$ are pre-defined locations on the rectified image $\image^r$, acting as constants and $\Delta_{\Theta^\prime}^{-1}$ is a constant matrix that depends only on $\Theta^\prime$. Let $\mathcal{P} = \{p_i \}_{i = 1, ..., N}$ be a grid of pixels on $\image$, then $p_i = \mathcal{T}\hat{p}^r_i$ describes the relationship between pixels on $\image$ and $\image^r$, where $\hat{p}^r_i$ is a uniform grid on $\image^r$ concatenated with some constants. Lastly, the rectified image is computed from the grid $\mathcal{P}$ via a bilinear sampler~\cite{Jaderberg2015}:
\begin{align}
I^r_{ci}=\sum_{n,m}^{H,W}I_{cnm}f(1-|p_{ix}-m|)f(1-|p_{iy}-n|),
\end{align}
where, for each channel $c$, $I^r_{ci}$ is the $i$-th pixel on $\image^r$, $I_{cnm}$ is pixel $(n,m)$ on $\image$, $p_{ix}$ and $p_{iy}$ are the x and y coordinates of the grid map $p_i$, and $f(\cdot)$ denotes the ReLU function.

\subsubsection{Connectionist Temporal Classification Decoder}

The CTC decoder uses an additional \textit{blank} token in $\classes$. Denote the set of all possible sequences of predicted classes with length $\frames$ as $\classes^\frames$, and let $\map$ be a many-to-one mapping which maps $\pi \in \classes^\frames$ to the decoded text $\ell$ by removing repeated predictions and \textit{blank} tokens. For instance, $\map($\texttt{-ff-l-yy--}$)=$ \texttt{fly} when $\frames=10$, where \texttt{-} denotes the \textit{blank} token. Given an input $\image$, the \textit{conditional probability} for $\ell$ conditioned on the network output $\network(\image)$ is defined as a sum over the preimage of $\ell$ under $\map$:
\begin{equation}
    \probability(\ell \mid \network(\image)) =\sum_{\pi\in\map^{-1}(\ell)}\probability(\pi \mid \network(\image)),\label{eq:ctc_1}
\end{equation}
where $\probability(\pi \mid \network(\image))=\prod{_{\frame=1}^\frames}\network_{\frame,\pi_\frame}(\image)$ is the product over probabilities of predicting label $\pi_\frame$ at frame $\frame$. The optimal decoded text $\ell^*$ is the maximum likelihood solution of  \ref{eq:ctc_1}, and we follow Graves \textit{et al.} \cite{Graves2006} to approximate the maximum likelihood solution by 
\begin{equation}
    \ell^* \approx \map(\arg\max_\pi (\pi \mid \network(\image))),\label{eq:ctc_2}
\end{equation}
meaning the most likely label for each frame is selected before applying $\map$ to recover $\ell^*$. The CTC decoder $\decoder$ is then defined as \ref{eq:ctc_2}, where $\decoder(\network(\image))=\ell^*$.

\subsubsection{Attention Decoder} 

The attention decoder relies on an LSTM to output the sequence of $\frames$ labels, where the inputs to each LSTM cell $t^\prime\in[1..\frames]$ are the attention weighted average over $\mathcal{H}$, the previous predicted label and the previous hidden state from cell $t^\prime-1$, as shown in  \ref{fig:attention_decoder}. Let $\network(\image)=\mathcal{H}=[h_1, ..., h_T]$ be the contextual features. For each $t^\prime\in[1..\frames]$, denote $s_{t^\prime}$ as the hidden state output of the $t^\prime$-th LSTM cell, and then $y_{t^\prime}=\sigma_{soft}(W_0 x_{t^\prime}+b_0)$ is the class prediction output for frame $t^\prime$ given the hidden states, where $W_0$ and  $b_0$ are trainable parameters and $\sigma_{soft}$ is the Softmax function.

To obtain the attention weights for the inputs to cell $t^\prime$, first define $e_{t^\prime,t}=a^\top\tanh(W s_{t^\prime-1}+V h_t +b)$, where $W,V,a,b$ are trainable parameters. Then, define
\begin{equation}
\alpha_{t^\prime,t}=\sigma_{soft}(e_{t^\prime,\cdot})=\frac{\exp(e_{t^\prime,t})}{\sum_{j=1}^T \exp(e_{t^\prime,j})}
\end{equation}
to be the attention weights. The attention operation then linearly combines contextual features $[h_1, ..., h_T]$ using the attention weights to obtain $g_{t^\prime}=\sum_{t=1}^T\alpha_{t^\prime,t}\cdot h_t$. The LSTM cell of the decoder is then recurrently updated by
\begin{equation}
s_{t^\prime}=\text{LSTM}([g_{t^\prime},h(y_{t^\prime-1})],s_{t^\prime-1}),
\end{equation}
where $y_{t^\prime-1}$ is the previous predicted label and $h(\cdot)$ is the one-hot embedding. This yields a sequence of predicted labels $y_{t^\prime}$, in which the \textit{end-of-sentence} token [s] indicates the end of the output word.




\subsection{Vision Transformers for STR}\label{sec:vit}

ViTSTR~\cite{Atienza2021VisionRecognition} is an adaptation of the standard Vision Transformer (ViT~\cite{Dosovitskiy2021}) to the STR task. ViT adopts the Transformer~\cite{vaswani2017attention} encoder originally designed for NLP tasks, where ViTSTR further extends ViT by modifying the prediction head to predict an ordered sequence of labels,  instead of a single label,  for classification.

The general architecture of ViTSTR is given in \ref{fig:vitstr}. The input image $\image\in\realNumber^{H \times W \times C}$ is cut into a sequence of $\frames$ patches shaped $P$$\times$$P$$\times$$C$. It is then flattened and converted via a linear projection to what we refer to as the sequence patch embedding. An extra learnable class embedding is prepended to the sequence of patch embeddings, where unique positional encodings are added to each patch embedding. The resulting patch embeddings are the input to the Transformer encoder, where $\frames$ features are extracted from the encoder instead of just one. A prediction head then projects these features into $\frames$ label predictions. This sequence of labels always start with an \textit{begin-of-sentence} token [GO], whereas the \textit{end-of-sentence} token [s] indicates the end of the output word. Unlike LSTM-based models, Transformers can predict the sequence of labels in parallel.

\section{STR-Cert}\label{sec:method}

We introduce STR-Cert, a polyhedral verification method based on DeepPoly~\cite{singh2019abstract}, and propose novel algorithms and polyhedral bounds that are necessary to certify STR models.

\subsection{Polyhedral Verification}
While several robustness certification methods have been proposed, as overviewed in \ref{sec:related}, STR-Cert utilizes polyhedral verification for its balance of scalability and precision, offering a sweet spot for STR certification. Specifically, we adopt the DeepPoly abstract domain~\cite{singh2019abstract}, which is a sub-polyhedral abstract domain that maintains lower and upper \textit{polyhedral bounds} and \textit{interval bounds} for each neuron. Formally, let $\mathcal{X} = \{x_1, x_2, ..., x_n\}$ be an ordered set of neurons in $\network$ such that the order complies with the order of the layers they belong to. For each neuron $x_j$, we define the interval bounds $l_j \leq x_j \leq u_j$ and the polyhedral bounds $\sum_{i<j}a^l_i\cdot x_i+b^l\leq x_j \leq \sum_{i<j}a^u_i\cdot x_i+b^u$, where $l_j,u_j,a^l_i,b^l,a^u_i,b^u\in\realNumber\cup\{-\infty,\infty\}$. To certify robustness, DeepPoly adopts \textit{backsubstitution} to bound the difference between neurons in the network logits output, i.e., whether $\network_{t,c}(\image^\prime)-\network_{t,c^\prime}(\image^\prime)>0 \quad \forall \image^\prime\in\advDomain_\image$, to certify no class change occurs under perturbation. It recursively substitutes target neurons with the polyhedral bounds of previous layers’ neurons until reaching the input neurons. We note that DeepPoly is algorithmically equivalent to CROWN~\cite{Zhang2018}, and we refer the reader to~\cite{singh2019abstract} for details of DeepPoly.

The polyhedral verification framework has been adopted to certifying FNCCs~\cite{singh2019abstract}, CNNs~\cite{Boopathy2019}, LSTMs~\cite{Ryou2020ScalableNetworks}, and the Softmax function~\cite{Wei2023ConvexVerification}. To certify STR models introduced in  \ref{sec:STR_arch}, we first derive novel polyhedral bounds for the network components not covered in the literature, namely TPS, patch embedding and positional encoding in  \ref{sec:stn} and  \ref{sec:patchEmb}, while other relevant polyhedral bounds are provided in  \ref{appendix:bounds}. Secondly, we bridge the gap between neural network predicted sequence of labels and the final predicted text. For attention decoder and ViTSTR models, the sequence of labels are directly used as the predicted text, meaning any change of label before the \textit{end-of-sentence} token guarantees a change in the predicted text. CTC decoder models, however, operate differently and we provide an algorithm to certify them in  \ref{sec:ctc}. Finally, since Softmax is a key component in attention decoder and ViTSTR, we propose novel polyhedral bounds for Softmax in  \ref{sec:refine} that refines existing bounds by considering the constraint that Softmax outputs sum to 1.

\subsection{TPS transformation}\label{sec:stn}

Recall from \ref{sec:transf} that the TPS transformation utilizes a localization CNN to produce the fiducial points $\Theta$, before computing matrix $\mathcal{T}$ and generating the grid of pixels $\mathcal{P}$ on $\image$, where $p_i=\mathcal{T}\hat{p}^r_i$. When $\image$ is under perturbation $\advDomain_\image$, polyhedral bounds for $\Theta$ can be computed by existing DeepPoly methods for CNNs, which can be directly applied to bound $\mathcal{T}$. Since $\hat{p}^r_i$ is a uniform grid on $\image^r$ that are unaffected by perturbations to $\image$, $\mathcal{P}$ is a linear transformation from $\mathcal{T}$, which can be adopted as the polyhedral bounds for $\mathcal{P}$.

\begin{figure}[t]
\begin{subfigure}[t]{0.5\textwidth}
    \centering
    \includegraphics[width=1\textwidth,frame]{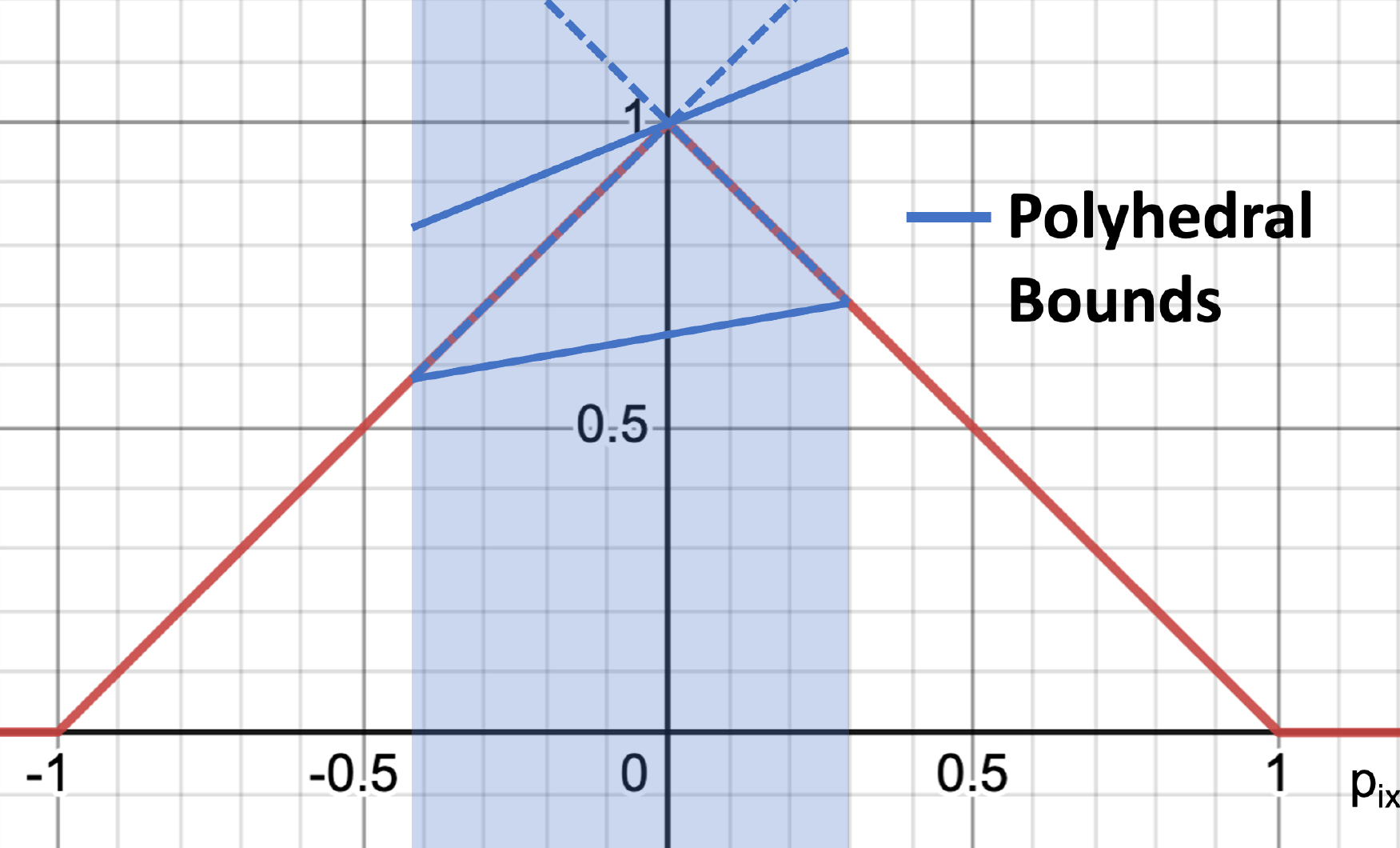}
    \caption{$l_{ix}\in[-1,0],u_{ix}\in[0,1]$}
    \label{fig:bilinear1}
\end{subfigure}
\begin{subfigure}[t]{0.5\textwidth}
    \centering
    \includegraphics[width=1\textwidth,frame]{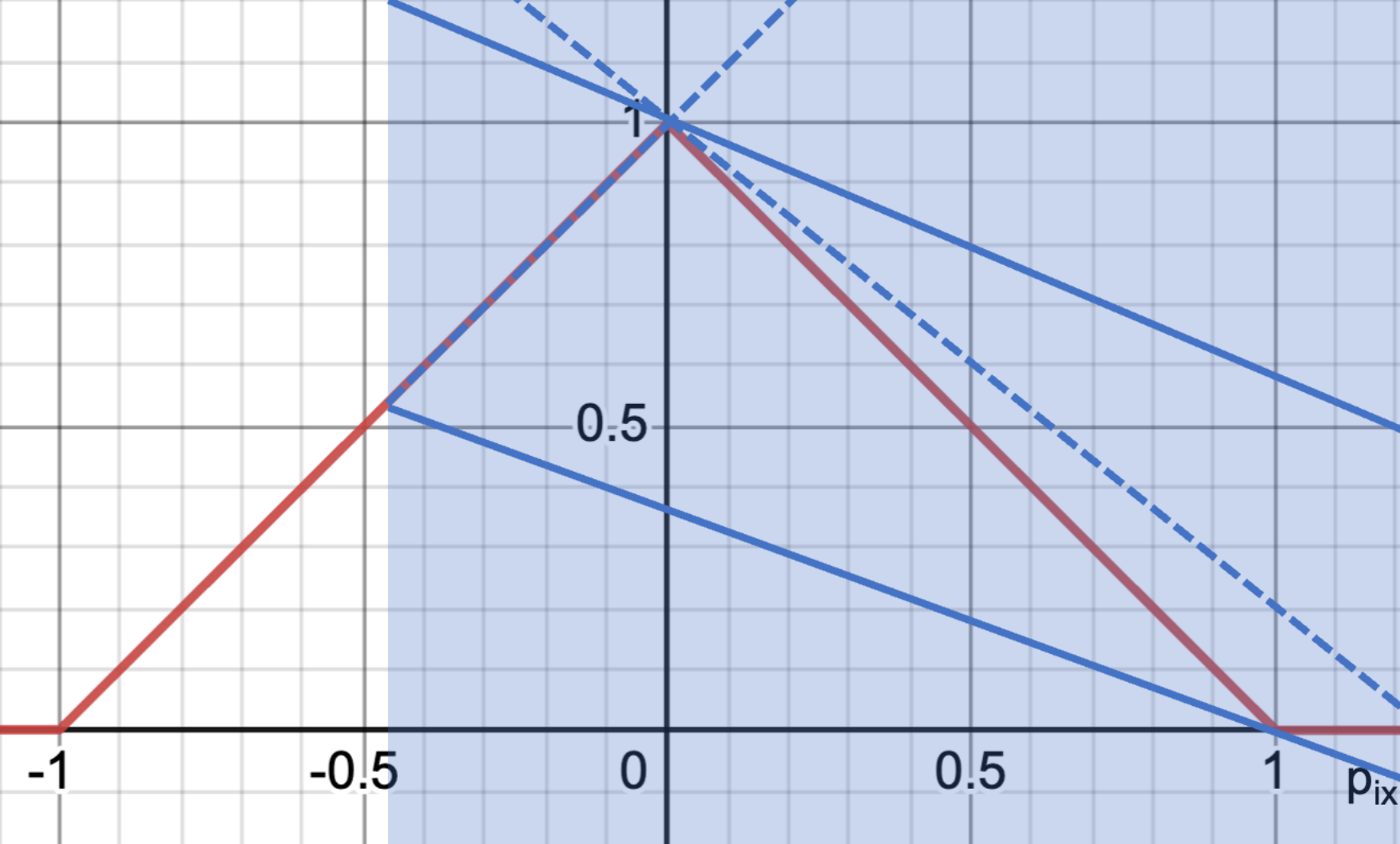}
    \caption{$l_{ix}\in[-1,0],u_{ix}\in[1,\infty]$}
    \label{fig:bilinear2}
\end{subfigure}
\caption{Polyhedral bounds for $f(1-|p_{ix}-m|)$ against $p_{ix}$ in the bilinear map of TPS.}
\vspace{-0.1cm}
\label{fig:bilinear}
\end{figure}

To derive polyhedral bounds for the rectified image $\image^r$, recall the bilinear map, where $f(\cdot)$ is the ReLU function:
\vspace{-0.1cm}
\begin{align}
I^r_{ci}=\sum_{n,m}^{H,W}I_{cnm}f(1-|p_{ix}-m|)f(1-|p_{iy}-n|).
\end{align}
Since $p_{ix}$, $p_{iy}$ and $I_{cnm}$ are all under perturbation, we first derive polyhedral bounds for $r_{ix}\coloneqq f(1-|p_{ix}-m|)$ from interval bounds $l_{ix}<p_{ix}<u_{ix}$. The function mapping $p_{ix}$ to $r_{ix}$ consists of four pieces of linear functions as shown in \ref{fig:bilinear}. Assume WLOG that $m=0$, then if $l_{ix},u_{ix}$ are within the left two or right two pieces of linear functions, i.e., $l_{ix},u_{ix}\in[-\infty,0]$ or $[0,\infty]$, the polyhedral bounds follows from ReLU (see \ref{appendix:relu}). If $l_{ix}\in[-1,0]$ and $u_{ix}\in[0,1]$, the polyhedral bounds are illustrated in \ref{fig:bilinear1}: 
\begin{equation}
\frac{-(u_{ix}+l_{ix})}{u_{ix}-l_{ix}}p_{ix}+\frac{2u_{ix}l_{ix}}{u_{ix}-l_{ix}}+1\leq r_{ix}\leq a^u_{1}p_{ix}+1,
\end{equation}
where $a^u_{1}\in[-1,1]$. If $l_{ix},u_{ix}$ are across three linear functions, first consider the case of \ref{fig:bilinear2}. The bounds are
\begin{equation}
\frac{-(1+l_{ix})}{1-l_{ix}}p_{ix}+\frac{2l_{ix}}{1-l_{ix}}+1\leq r_{ix}\leq a^u_{2}p_{ix}+1,
\end{equation}
where $a^u_{2}\in[-1/u_{ix},1]$. The bounds for the other case of $l_{ix}\in[-\infty,-1],u_{ix}\in[0,1]$ follow from symmetry. Lastly, if $l_{ix},u_{ix}$ are across all four linear functions, the bounds are $0\leq r_{ix}\leq a^u_{3}p_{ix}+1$, where $a^u_{3}\in[-1/u_{ix},-1/l_{ix}]$. Bounds for $r_{iy}\coloneqq f(1-|p_{iy}-n|)$ can be derived similarly. In practice, $a^u_{1}$, $a^u_{2}$ and $a^u_{3}$ are chosen such that the area bounded by the interval and polyhedral bounds is minimized. Detailed bounds for all cases are provided in \ref{appendix:bilinear}. With the polyhedral bounds of $I_{cnm}$, $r_{ix}$ and $r_{iy}$, the final polyhedral bounds for $\image^r$ can be computed via addition and multiplication bounds (see \ref{appendix:multi}).

\subsection{Patch Embedding and Positional Encoding}\label{sec:patchEmb}

Patch embedding and positional encoding are two crucial components of ViTSTR as introduced in \ref{sec:vit}. Recall that patch embedding produces a sequence of $\frames$ patches before flattened into vectors of dimension $P^2C$ and linearly projected into a embedding of size $D$. The composition of patching $\phi_1: \mathbb{R}^{H \times W \times C} \rightarrow \mathbb{R}^{N \times P^2C}$ and linear projection $\phi_2:\mathbb{R}^{N \times P^2C}\rightarrow \mathbb{R}^{N \times D}$ is equivalent to a convolution operation with kernel size $(P\times P)$, stride size $P$, input channels $C$ and output channels $D$. The kernel weights is obtained by reshaping the linear projection weights from shape $(P^2C\times D)$ to $(D\times C\times P\times P)$. The patch embedding can thus be certified as a convolution layer.

We denote the outputs from the patch embedding as $v^p \in \mathbb{R}^{N \times D}$. A unique positional encoding of the same dimension $D$ is added to each patch (for details see~\cite{Atienza2021VisionRecognition}). Let $e^p \in \mathbb{R}^{N \times D}$ be the positional encoding matrix for the whole image, then the positional encoding layer corresponds to $v^p + e^p$. Since $e^p$ is constant with respect to perturbations, positional encoding can be certified as a linear layer with the identify matrix as weight and $e^p$ as bias.

\subsection{CTC Decoder Certification}\label{sec:ctc}

\begin{algorithm}[tb]
\DontPrintSemicolon
\SetNlSty{textbf}{}{:}
\caption{CTC Decoder Certification}
\label{alg:ctc}
\BlankLine
\SetKw{KwIn}{in}
\SetKwInOut{Input}{Input}
\Input{CTC decoder $\decoder$, input $\image$, network $\network$, adversarial region $\advDomain_\image$, verifier $V$}
  {
  \nl $\ell^*\gets\decoder(\network(\image))$\\
  \nl \For{$\frame\gets\frames$}{
  \nl $l_t \gets\text{argmax}_k (\network_{t,k}(\image))$\\
  \nl $M_t\gets[l_t]$\\
  \nl \For{$\class\gets \classes$}{
  \nl \uIf{$V(\mathbb{S}_\image, \network_{t,l_t}>\network_{t,c})$=unsafe, $l_t\neq c$}{
  \nl $M_t.append(c)$\\}}
  \nl \uIf{$|{M_t}|>3$}{
  \nl \Return Unsafe 
  }}
  \nl \For{$\pi \gets (M_1\times M_2\times...\times M_T)$}{
  \nl \uIf {$\map(\pi)\neq\ell^*$}{
  \nl \Return Unsafe\\
  }}
  \nl \Return Safe }
  \vspace{-0.1cm}
\end{algorithm}

Recall that $\network_{\frame,\class}(I)$ is the logits output for frame $t$ and class $\class$. To certify the output sequence of the CTC decoder $\decoder$, we need to check that, for the perturbed input domain $\mathbb{S}_\image$, all the combinations of possible predictions from each frame $t$ still produce the ground truth sequence after applying reduction mapping $\map$. Algorithm~ \ref{alg:ctc} outlines the procedure for CTC decoder certification. At line 6, $V(\mathbb{S}_\image, \network_{t,l_t}>\network_{t,c})$ returns unsafe if these exist $\image^\prime\in\mathbb{S}_\image$ such that $\network_{t,l_t}(\image^\prime)\leq\network_{t,c}(\image^\prime)$. Therefore, $M_t$ is the set of all possible classifications for frame $t$ under perturbation. To justify line 8-9, if there exists $t$ such that $|M_t|>3$, let $l_t$ be the true prediction label for frame $t$. Then, there must exist $c_t\in M_t$ such that $c_t\neq l_{t-1},l_{t}$ and $l_{t+1}$ by the pigeon hole principle. This means $c_t$ will not be a repeated label and $\map$ will produce a different sequence to the ground truth, failing the certification. If $|M_t|\leq 3$ for all $t$, we check all possible sequences $\pi\in (M_1\times M_2\times ...\times M_T)$ for text changes after applying $\map$ at line 10-12. If all possible sequences are safe, the CTC decoder model is then certified to be safe under $\advDomain_\image$. Note that the worst-case complexity for \ref{alg:ctc} is $\mathcal{O}(3^{|T|})$, but without line 8-9, it will increase to $\mathcal{O}(|\classes^\frames|)$, where often $|\classes|>40$. However, we find that, in practice, $|M_t|$ increases monotonically with $t$, and usually quite dramatically because LSTM certification loses precision as the network is unfolded. This means Algorithm~ \ref{alg:ctc} typically terminates at line 9, and if not, the search space of $(M_1\times M_2\times ...\times M_T)$ is in the order of $4!$, which is constant time.

\subsection{Refining Softmax Bounds}\label{sec:refine}

Softmax is an important function in both ViTSTR and the attention decoder. Existing Softmax polyhedral bounds~\cite{Wei2023ConvexVerification,Shi2020RobustnessTransformers} only consider element-wise bounds between the inputs and outputs of Softmax, and they fail to consider the crucial constraint that the Softmax output sum up to 1. Let the Softmax output neurons $x_i$, $i\in [1..N]$ have interval bounds $l_i\leq x_i \leq u_i$, respectively. We aim to refine Softmax by introducing a novel polyhedral transformation of the existing Softmax polyhedral bounds that incorporates the additional constraint $\sum^N_{i=1}{x_i}=1$. This constraint 
unfortunately cannot be described by a linear combination of previous neurons. Instead of over-approximating lower and upper polyhedral bounds, we find the minimum relaxation to the constraints such that a linear transformation can exactly describe it. By removing the interval bound constraint for a single neuron $x_k$, the following constraints
\vspace{-0.2cm}
\begin{align}
    l_i\leq x^r_i \leq u_i \quad \forall i\neq k \quad\text{ where } \sum^N_{i=1}{x^r_i}=1,\label{eq:relax_constraint}
\end{align}
can be exactly satisfied by the refined neurons $x_i^r$:
\vspace{-0.1cm}
\begin{align}
    x^r_i=x_i\text{ for } i\neq k\text{ and }x^r_k=1-\sum_{i\neq k}{x_i}.\label{eq:refined_var}
\end{align}
In this case, the polyhedral bounds are this linear transformation, making the polyhedral transformation exact. Next, we discuss how to choose $k$ such that the constraint of \ref{eq:relax_constraint} is the tightest. From \ref{eq:refined_var} and \ref{eq:relax_constraint}, we can deduce implied upper and lower bounds for $x^r_k$:
\vspace{-0.05cm}
\begin{align}
    1-\sum_{i\neq k}u_i\leq x^r_k=1-\sum_{i\neq k}{x_i}=\leq 1-\sum_{i\neq k}l_i,\label{eq:ineq}
\end{align}
where the tightness of these implied bounds for $x^r_k$ is
\vspace{-0.05cm}
\begin{align}
(1-\sum_{i\neq k}l_i)-(1-\sum_{i\neq k}u_i)=\sum_{i\neq k}(u_i-l_i).
\end{align}
Therefore, by choosing $k=\argmax_{i} (u_i-l_i)$ with the loosest bounds, the implied bounds for $x^r_k$ will be the tightest, ensuring it is a minimum relaxation from the original constraints. Note that, when $1-\sum_{i\neq k}l_i\leq u_k$ and $1-\sum_{i\neq k}u_i\geq l_k$, the inequality $l_k\leq x^r_k \leq u_k$ can be inferred from \ref{eq:ineq} and we can impose the $\sum^N_{i=1}{x_i}=1$ constraint without any relaxation.

\section{Experiments}

We present STR-Cert\footnote{Will be made open-source with the final version of this paper.} by implementing \ref{sec:method} and extending DeepPoly~\cite{singh2019abstract}. We also adopt Prover~\cite{Ryou2020ScalableNetworks} for certifying LSTMs and the element-wise Softmax bounds derived by Wei \textit{et al.}~\cite{Wei2023ConvexVerification}. We evaluate the performance and precision of STR-Cert on a range of STR networks and datasets, while providing insights into the comparisons between STR architectures and connections between adversarial training, prediction confidence, and robustness.

\subsection{Datasets, Models and Training}

We adopt the training setup of Baek \textit{et al.}~\cite{Baek2019} with the PyTorch~\cite{Paszke2019PyTorch:Library} library. For all architectures, the models are trained for 100000 iterations on the synthetic MJSynth~\cite{Jaderberg2014} and SynthText~\cite{Gupta2016} datasets and validated on the training datasets of IIIT5K~\cite{Mishra2012}, IC13 Born-Digital Images~\cite{Karatzas2013}, IC15 Focused Scene Text~\cite{Karatzas2015}, SVT~\cite{Wang2011}, SVTP~\cite{Phan2013} and CUTE80~\cite{Risnumawan2014}. We use AdamW~\cite{Kingma2015} optimizer with a cosine decay scheduler, while also deploying PGD adversarial training~\cite{Madry2018}, discussed further in \ref{sec:adv_training}. The attention decoder model utilizes TPS~\cite{Shi2016} as the transformation module, a 5 layer CNN-ReLU network as the feature extractor, an LSTM as the sequence model, and the attention decoder~\cite{Wojna2017Attention-BasedImagery}. The CTC decoder model uses the same transformation, feature extractor, and sequence model architectures. For the Vision Transformer model, we use the architecture in ViTSTR~\cite{Atienza2021VisionRecognition} with 5 layers. The pre-trained ViTSTR models~\cite{Atienza2021VisionRecognition} are unfortunately too large and beyond the scope of this work. The CTC decoder and the attention decoder models both have around 500K parameters, whereas the ViTSTR have around 700K parameters. Details on architectures and training can be found in \ref{appendix:arch}.

\subsection{Robustness Certification}
\begin{figure*}[t]
\centering
\begin{subfigure}[t]
{0.33\textwidth}
\centering\includegraphics[width=1\textwidth]{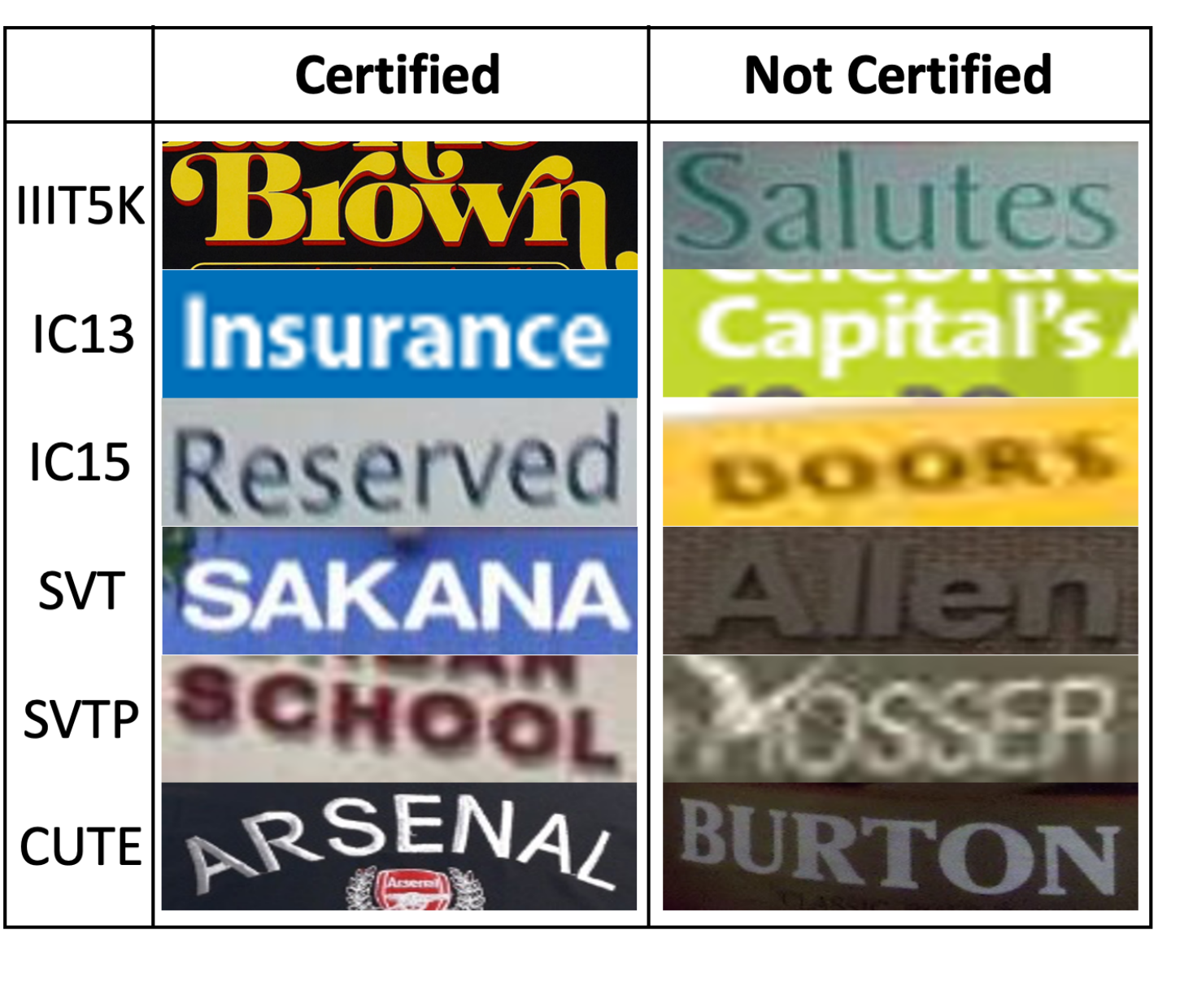}
\captionsetup{width=.95\linewidth}
\caption{Examples from each dataset that are certified and not certified against adversarial attacks.}
    \label{fig:sample_photos}
\end{subfigure}
\begin{subfigure}[t]{0.32\textwidth}
\centering
\captionsetup{width=.95\linewidth}
\includegraphics[width=1\textwidth]{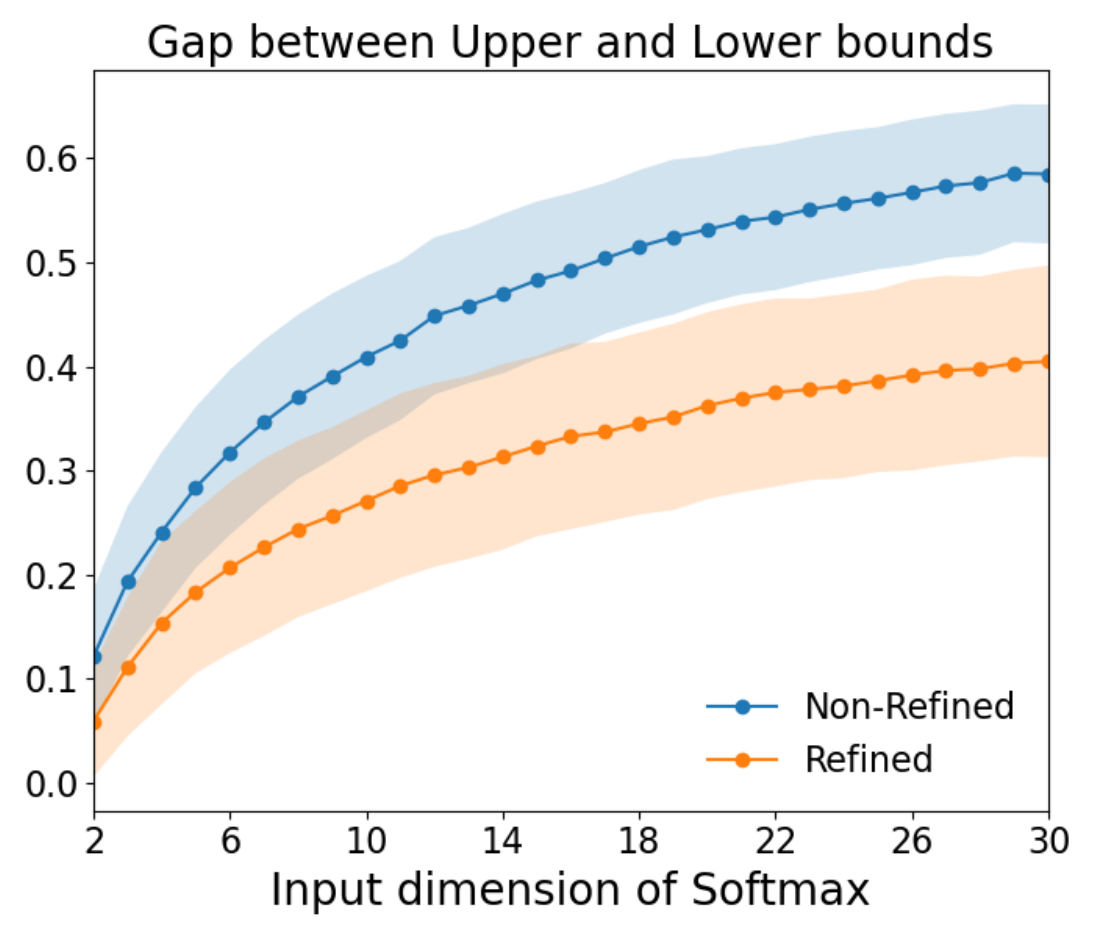}
\caption{Tightness of bounds with and without Softmax refinement against input dimensions.}
\label{fig:softmax}
\end{subfigure}
\begin{subfigure}[t]{0.32\textwidth}
\centering\captionsetup{width=.9\linewidth}\includegraphics[width=1\textwidth]{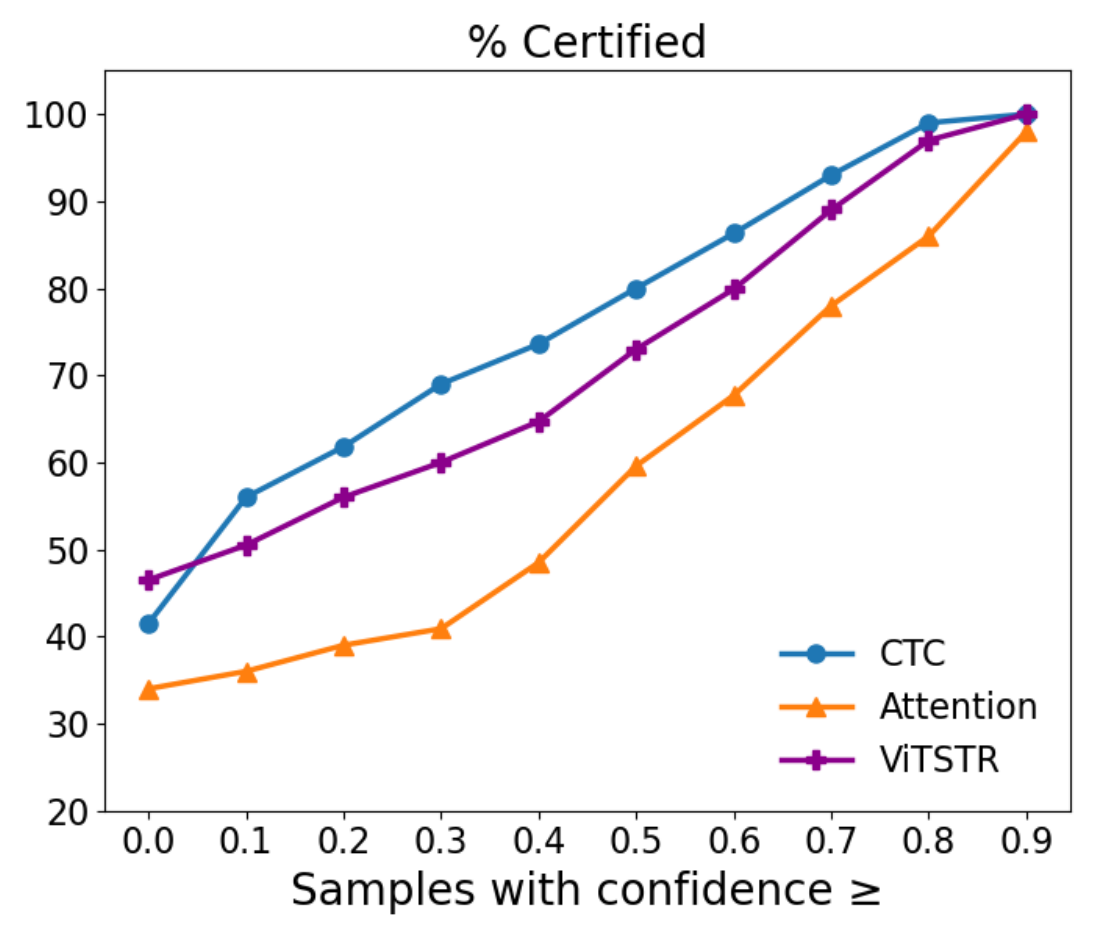}
\caption{\% certified for samples with prediction confidence above various thresholds.}
\label{fig:confidence}
\end{subfigure}
\vspace{-0.1cm}
\caption{Example images and certification results for Softmax bounds and prediction confidence.}
\vspace{-0.1cm}
\end{figure*}

\begin{table*}[t]
    \centering
    \tiny
    \tabcolsep=0.14cm
    \begin{tabular}{c|ccc|ccc|cccc}
    \toprule
    Model&\multicolumn{3}{c}{CTC decoder} &\multicolumn{3}{c}{Attention decoder}&\multicolumn{4}{c}{ViTSTR} \\
    \cmidrule(lr){2-4}\cmidrule(lr){5-7}\cmidrule(lr){8-11}
    Datasets&$\epsilon=.001$ & $\epsilon=.003$ & $\epsilon=.005$ &$\epsilon=.001$ & $\epsilon=.003$ & $\epsilon=.005$ &$\epsilon=.001$ & $\epsilon=.003$ & $\epsilon=.005$ & $\epsilon=.01$\\
    \midrule
    IIIT5K & 98.5\% & 76.5\% & 48.5\% & 91.0\% & 68.0\% & 39.0\% & 97.5\% & 75.0\% & 57.5\% & 24.0\%\\
    IC13   & 99.5\% & 89.5\% & 59.0\% & 95.0\% & 82.0\% & 52.0\% & 98.5\% & 87.0\% & 74.5\% & 41.5\%\\
    IC15   & 95.5\% & 56.5\% & 18.0\% & 89.5\% & 35.0\% & 11.0\% & 95.5\% & 58.5\% & 41.0\% & 11.5\%\\
    SVT    & 95.0\% & 68.0\% & 34.0\% & 88.0\% & 42.5\% & 19.0\% & 97.5\% & 65.5\% & 58.5\% & 29.5\%\\
    SVTP   & 94.5\% & 62.5\% & 36.5\% & 88.5\% & 58.0\% & 21.5\% & 96.0\% & 61.5\% & 43.0\% & 21.5\%\\
    CUTE   & 97.5\% & 83.0\% & 41.0\% & 92.0\% & 73.0\% & 33.0\% & 96.5\% & 79.5\% & 58.0\% & 27.0\%\\
    \bottomrule
    \end{tabular}
        \caption{\% certified in the first 200 correctly classified instances for CTC decoder, attention decoder and ViTSTR model for 6 datasets.}
    \label{tab:verifiability}
\vspace{-0.1cm}
\end{table*}
\begin{figure*}[tb]
\centering
\begin{subfigure}[t]
{0.33\textwidth}
\centering
\includegraphics[width=1\textwidth]{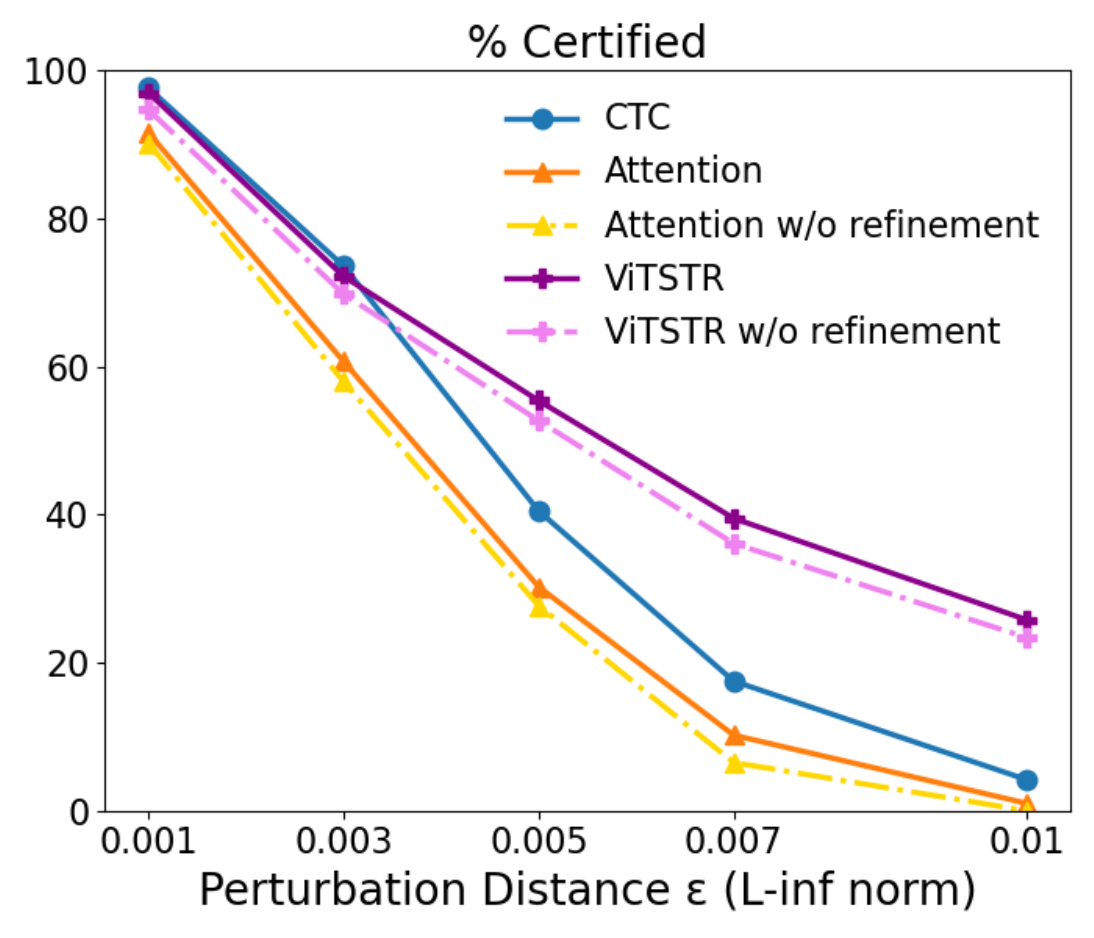}
    \caption{\% certified against perturbation distances.}
    \label{fig:summary}
\end{subfigure}
\begin{subfigure}[t]{0.32\textwidth}
\centering\captionsetup{width=0.9\linewidth}\includegraphics[width=1\textwidth]{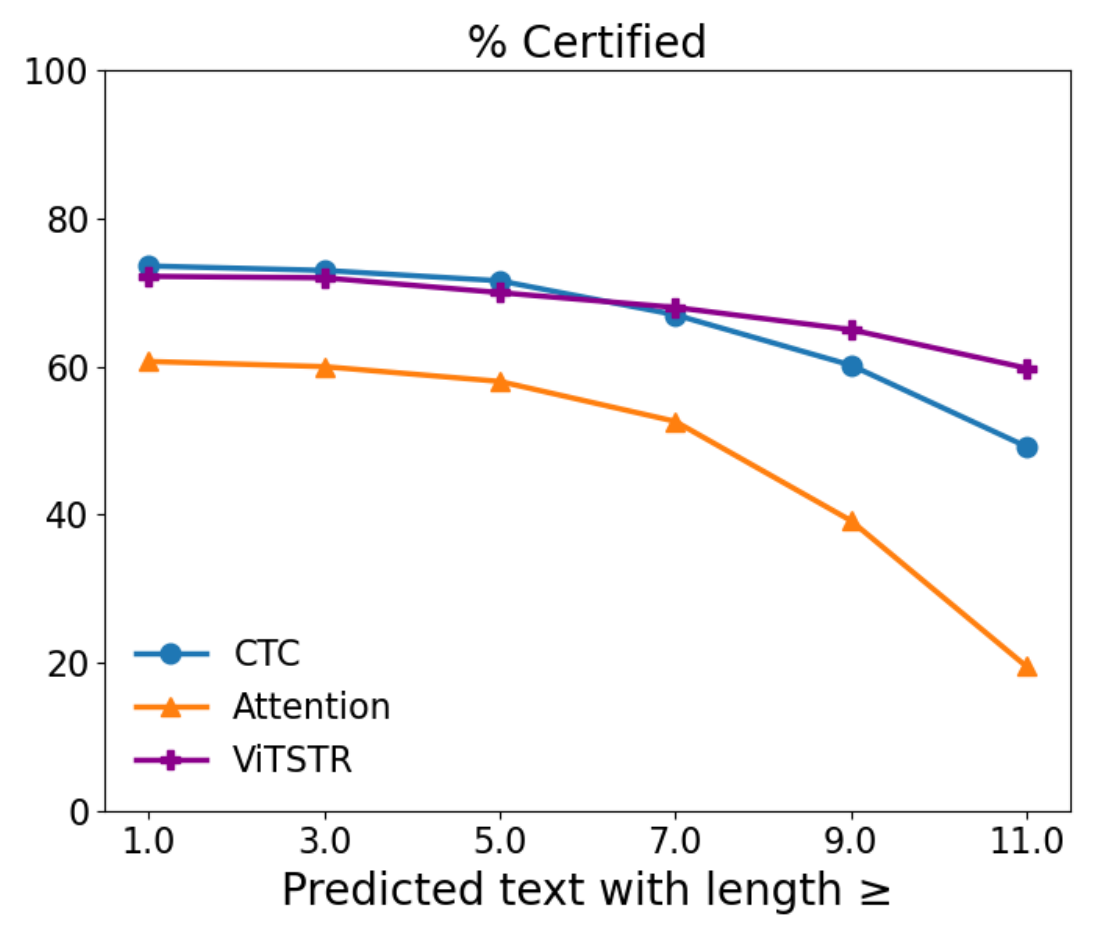}
\caption{\% certified for samples with predicted text longer then various lengths.}
    \label{fig:frames}
\end{subfigure}
\begin{subfigure}[t]{0.32\textwidth}
\centering\captionsetup{width=.9\linewidth}
\includegraphics[width=1\textwidth]{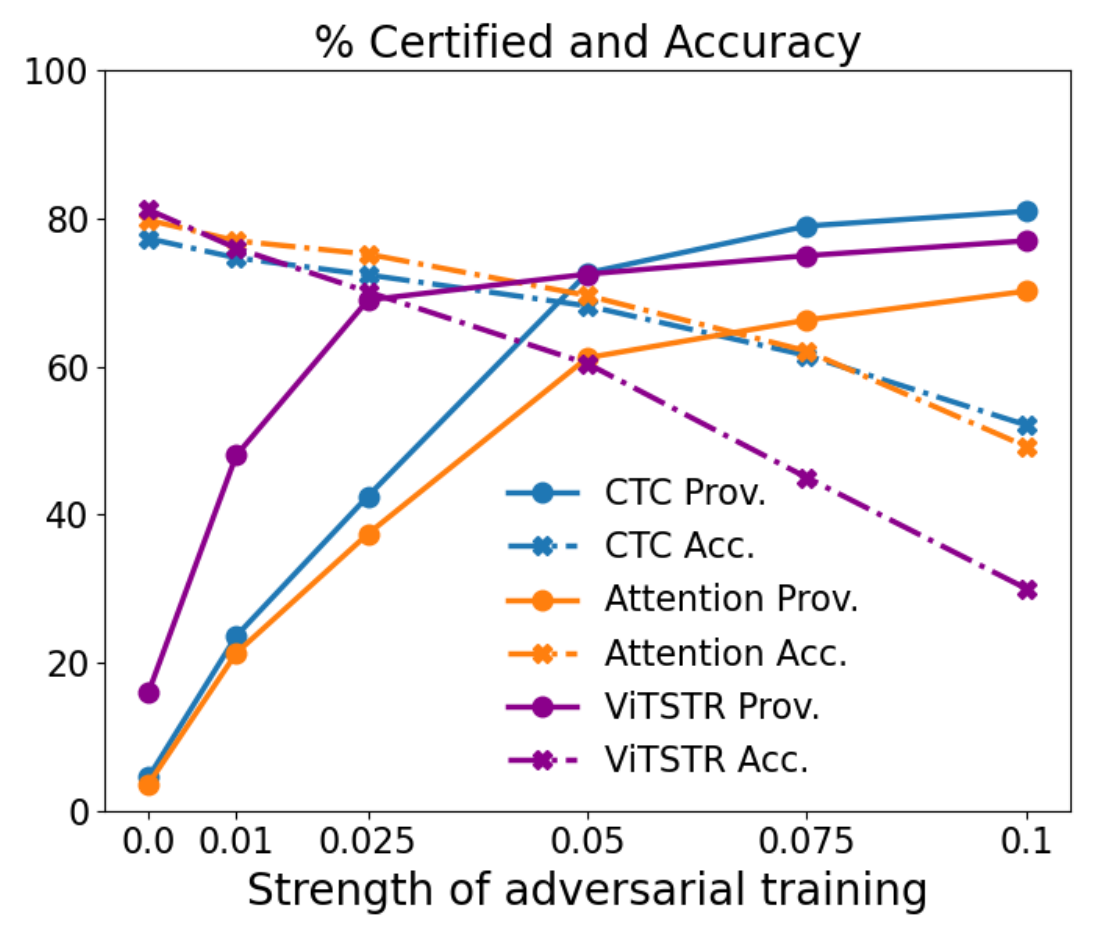}
\caption{\% certified and accuracy against the strength of PGD adversarial training.}
\label{fig:adv_training}
\end{subfigure}
\vspace{-0.1cm}
\caption{Certification results with analysis against text length and adversarial training strength.}
\vspace{-0.1cm}
\end{figure*}
We now provide results on the certified robustness for the CTC decoder, attention decoder, and ViTSTR model on the test datasets of IIIT5K~\cite{Mishra2012}, IC13~\cite{Karatzas2013}, IC15~\cite{Karatzas2015}, SVT~\cite{Wang2011}, SVTP~\cite{Phan2013} and CUTE80~\cite{Risnumawan2014}. For each model, we examine the percentage of correctly classified samples that can be certified to remain correctly classified under perturbation, which we refer to as \emph{percentage certified}. We certify the first 200 correctly classified samples in each dataset for varying perturbation budget, i.e., the maximum perturbation distance, of $\epsilon=0.001$ up to $0.01$ under $\linfty$ norm. The average percentage certified 
for all architectures with and without Softmax refinement (in \ref{sec:refine}) are illustrated against perturbation budgets in \ref{fig:summary}, and detailed results for certification with Softmax refinement are shown in \ref{tab:verifiability}. Note that the CTC decoder model does not have a Softmax layer, and an ablation study for Softmax refinement is provided in \ref{sec:ablation}. We observe that the percentage certified for the CTC and attention decoder models drops more steeply for increasing perturbation budget $\epsilon$ than for the ViTSTR model. This is because each LSTM recurrent cell involves multiplication of two non-linear activated neurons, which is difficult to certify with high precision. Moreover, existing certifiers for LSTMs unfold the recurrent operation, which significantly increases the depth of the network on which the loss of precision compounds. In addition, harder datasets such as IC15 are consistently less robust across all model architectures and perturbation budgets. Fig.~ \ref{fig:sample_photos} includes examples of certified and uncertified samples, and additional certification results for STR models with different number of layers are provided in \ref{appendix:exp}, where similar trends can be observed.

Scalability with respect to the length of predicted text is particularly interesting. CTC and attention decoder models rely on LSTMs to recurrently predict the sequence of labels, which means the loss of precision compounds as the length of text increases. In practice, as shown in \ref{fig:frames}, percentage certified drops significantly for the two LSTM-based models as the predicted text grow longer then 9, especially for the attention-decoder model, which contains two layers of LSTM and struggles to certify predicted text with length 11 or more. ViTSTR, however, shows a smaller drop in percentage certified because it predicts in parallel, where longer text does not increase the depth of certification.

The average certification runtime per sample is 49s for CTC decoder model; 92s for attention decoder model; and 14s for ViTSTR. The discrepancy between runtime is mainly due to the certification of LSTM layers, as the CTC decoder and attention decoder models include one and two LSTM layers, respectively. If we take into consideration that large-scale LSTM-based STR models in practice usually include multiple layers of Bi-LSTM, it becomes extremely challenging, if not infeasible, to certify them with current methods. Vision Transformers, however, seem to be a better choice in terms of certification scalability.

\subsection{Effect of Adversarial Training}\label{sec:adv_training}

To investigate the effect of adversarial training on percentage certified, we train various models in all three STR architectures with different strengths of adversarial training, i.e., the maximum perturbation budget in PGD~\cite{Madry2018}, with 10 steps. For CTC and attention decoder models, standard PGD adversarial training is adopted to train models from scratch. For the ViTSTR model, we follow the adversarial training setup of ViT~\cite{Mo2022} and naturally train the model for 20000 epochs before training adversarially, which improves the training stability and adversarial accuracy.

The average percentage certified and accuracy across the test datasets for STR architectures are shown in \ref{fig:adv_training}, where the adversarial training strength ranges from 0 (natural training) to 0.1 and the samples are certified against perturbation budget $\epsilon=0.003$. We observe that, as expected, there exists a trade-off between the accuracy of the model and the percentage certified. However, it is interesting to note that the optimal trade-off points vary for the different architectures. For CTC and attention decoder models, as the PGD strength rises above 0.05, the accuracy begins to drop markedly whilst the percentage certified only increases marginally. For ViTSTR, the optimal strength is, however, around 0.025 since the percentage certified stagnates with higher strength while the accuracy plummets. Nevertheless, the experiments confirm that adversarial training significantly boosts percentage certified.

\subsection{Ablation on Softmax Refinement}\label{sec:ablation}


To demonstrate effectiveness of our Softmax refinement (\ref{sec:refine}), we first directly compare the tightness of the output neuron's bounds with and without Softmax refinement on a synthetic neural network. A feed-forward network with 2 hidden layers and ReLU activations, followed by the Softmax layer and a fully-connected layer that outputs a single scalar is used. We compare the gap between the upper and lower bounds of the final output neuron under $\epsilon=0.1$ perturbation, with varying input dimensions to the Softmax function from 2 to 30. For each input dimension, we certify 1000 models with random parameters for the final layer and random inputs. The mean plus and minus one standard deviation of the gaps between bounds are shown in \ref{fig:softmax}, where the Softmax refinement provides considerably tighter bounds, even when the input dimensions for Softmax are high, as in the case of Vision Transformers.

On large-scale STR models, we demonstrate the improvement to percentage certified from Softmax refinement in \ref{fig:summary}. This improvement is present in all datasets for the attention decoder model and ViTSTR. In practice, we never observe an instance that the Softmax refinement harms the percentage certified.


\subsection{Prediction Confidence and Certification}

Since polyhedral verification provides bounds on the logits output of the network, samples with high prediction confidence, i.e., the product of confidence over the sequence of predictions, should be easier to certify. In \ref{fig:confidence}, percentage certified for samples with prediction confidence above various thresholds are plotted. When the lower confidence samples are filtered out, percentage certified dramatically increases, up to near 100\% when only high confidence samples remain. For the CTC decoder model, there are certain frames $t$ that can predict different labels without changing the decoded text. Those frames usually have lower prediction confidence, but this does not necessarily imply a lack of robustness. This is evident in \ref{fig:confidence} since the percentage certified for the CTC decoder model actually exceeds that of other architectures when the confidence threshold $\geq 0.1$.

\section{Conclusion}

We developed STR-Cert, the first robustness certification method for STR models and Vision Transformers, with novel algorithms and polyhedral bounds. We certified and compared three STR model architectures, where we demonstrated scalability issues of LSTM-based models and elucidated the benefits of Vision Transformers. 
Future work includes studying the robustness of pre-trained ViTSTR models in relation to its training dataset, extending the method to incorporate branch and bound, GPU parallelization, and certifying perturbations in other $L_p$ norms.

\bibliographystyle{plain}
\bibliography{reference} 

\begin{thebibliography}{10}

\bibitem{akintunde2019verification}
Michael~E Akintunde, Andreea Kevorchian, Alessio Lomuscio, and Edoardo Pirovano.
\newblock {Verification of RNN-based neural agent-environment systems}.
\newblock In {\em Proceedings of the AAAI Conference on Artificial Intelligence}, volume~33, pages 6006--6013, 2019.

\bibitem{Atienza2021VisionRecognition}
Rowel Atienza.
\newblock Vision transformer for fast and efficient scene text recognition.
\newblock {\em Proceedings of the International Conference on Document Analysis and Recognition, ICDAR}, 12821 LNCS:319--334, 5 2021.

\bibitem{Bacci2021}
E~Bacci, M~Giacobbe, and D~Parker.
\newblock Verifying reinforcement learning up to infinity.
\newblock {\em Proceedings of the International Joint Conference on Artificial Intelligence}, 2021.

\bibitem{Baek2019}
Jeonghun Baek, Geewook Kim, Junyeop Lee, Sungrae Park, Dongyoon Han, Sangdoo Yun, Seong~Joon Oh, and Hwalsuk Lee.
\newblock What is wrong with scene text recognition model comparisons? dataset and model analysis.
\newblock {\em Proceedings of the IEEE International Conference on Computer Vision}, 2019-October:4714--4722, 4 2019.

\bibitem{BonaertETHZurich2021FastTransformers}
Gregory Bonaert ETH~Zurich, Dimitar~I Dimitrov ETH~Zurich, Maximilian Baader ETH~Zurich, and Martin Vechev ETH~Zurich.
\newblock Fast and precise certification of transformers.
\newblock {\em International Conference on Programming Language Design and Implementation}, 2021.

\bibitem{Boopathy2019}
Akhilan Boopathy, Tsui-Wei Weng, Pin-Yu Chen, Sijia Liu, and Luca Daniel.
\newblock {CNN-Cert: An efficient framework for certifying robustness of convolutional neural networks}.
\newblock {\em Proceedings of the AAAI Conference on Artificial Intelligence}, 33:3240--3247, 2019.

\bibitem{Cohen2019CertifiedSmoothing}
Jeremy Cohen, Elan Rosenfeld, and J.~Zico Kolter.
\newblock Certified adversarial robustness via randomized smoothing.
\newblock {\em 36th International Conference on Machine Learning, ICML 2019}, 2019-June:2323--2356, 2 2019.

\bibitem{Dosovitskiy2021}
Alexey Dosovitskiy, Lucas Beyer, Alexander Kolesnikov, Dirk Weissenborn, Xiaohua Zhai, Thomas Unterthiner, Mostafa Dehghani, Matthias Minderer, Georg Heigold, Sylvain Gelly, Jakob Uszkoreit, and Neil Houlsby.
\newblock An image is worth 16x16 words: Transformers for image recognition at scale.
\newblock {\em ICLR 2021 - 9th International Conference on Learning Representations}, 10 2021.

\bibitem{dvijotham2018dual}
Krishnamurthy Dvijotham, Robert Stanforth, Sven Gowal, Timothy~A Mann, and Pushmeet Kohli.
\newblock A dual approach to scalable verification of deep networks.
\newblock In {\em UAI}, volume~1, page~3, 2018.

\bibitem{Fischer2021}
Marc Fischer, Maximilian Baader, and Martin Vechev.
\newblock Scalable certified segmentation via randomized smoothing.
\newblock {\em Proceedings of the International Conference on Machine Learning}, 139:3340--3351, 7 2021.

\bibitem{Goodfellow2015}
Ian~J. Goodfellow, Jonathon Shlens, and Christian Szegedy.
\newblock {Explaining and harnessing adversarial examples}.
\newblock In {\em 3rd International Conference on Learning Representations, ICLR}, 2015.

\bibitem{Graves2006}
Alex Graves, Santiago Fern{\'{a}}ndez, Faustino Gomez, and Jürgen Schmidhuber.
\newblock {Connectionist temporal classification: Labelling unsegmented sequence data with recurrent neural networks}.
\newblock In {\em ACM International Conference Proceeding Series}, volume 148, pages 369--376, 2006.

\bibitem{Gupta2016}
Ankush Gupta, Andrea Vedaldi, and Andrew Zisserman.
\newblock Synthetic data for text localisation in natural images.
\newblock {\em Proceedings of the IEEE Conference on Computer Vision and Pattern Recognition}, 2016-December:2315--2324, 4 2016.

\bibitem{Jaderberg2014}
Max Jaderberg, Karen Simonyan, Andrea Vedaldi, and Andrew Zisserman.
\newblock Synthetic data and artificial neural networks for natural scene text recognition.
\newblock {\em In Workshop on Deep Learning, NIPS}, 6 2014.

\bibitem{Jaderberg2015}
Max Jaderberg, Karen Simonyan, Andrew Zisserman, and Koray Kavukcuoglu.
\newblock Spatial transformer networks.
\newblock {\em Advances in Neural Information Processing Systems}, 2015-January:2017--2025, 6 2015.

\bibitem{Jordan2020}
Matt Jordan and Alexandros~G Dimakis.
\newblock {Exactly computing the local Lipschitz constant of ReLU networks}.
\newblock {\em Conference on Neural Information Processing Systems}, 2020.

\bibitem{Karatzas2015}
Dimosthenis Karatzas, Lluis Gomez-Bigorda, Anguelos Nicolaou, Suman Ghosh, Andrew Bagdanov, Masakazu Iwamura, Jiri Matas, Lukas Neumann, Vijay~Ramaseshan Chandrasekhar, Shijian Lu, Faisal Shafait, Seiichi Uchida, and Ernest Valveny.
\newblock {ICDAR 2015 competition on robust reading}.
\newblock {\em Proceedings of the International Conference on Document Analysis and Recognition, ICDAR}, 2015-November:1156--1160, 11 2015.

\bibitem{Karatzas2013}
Dimosthenis Karatzas, Faisal Shafait, Seiichi Uchida, Masakazu Iwamura, Lluis Gomez~I. Bigorda, Sergi~Robles Mestre, Joan Mas, David~Fernandez Mota, Jon~Almazan Almazan, and Lluis~Pere De~Las~Heras.
\newblock {ICDAR 2013 robust reading competition}.
\newblock In {\em Proceedings of the International Conference on Document Analysis and Recognition, ICDAR}, 2013.

\bibitem{Katz2017Reluplex:Networks}
Guy Katz, Clark Barrett, David~L. Dill, Kyle Julian, and Mykel~J. Kochenderfer.
\newblock {Reluplex: An efficient SMT solver for verifying deep neural networks}.
\newblock {\em Computer Aided Verification}, 10426 LNCS:97--117, 2017.

\bibitem{Kingma2015}
Diederik~P. Kingma and Jimmy~Lei Ba.
\newblock {Adam: A method for stochastic optimization}.
\newblock In {\em 3rd International Conference on Learning Representations, ICLR}, 2015.

\bibitem{Ko2019}
Ching~Yun Ko, Zhaoyang Lyu, Tsui~Wei Weng, Luca Daniel, Ngai Wong, and Dahua Lin.
\newblock {POPQORN: Quantifying robustness of recurrent neural networks}.
\newblock In {\em 36th International Conference on Machine Learning, ICML}, volume 2019-June, pages 6031--6087, 2019.

\bibitem{Li2023}
Linyi Li, Tao Xie, and Bo~Li.
\newblock {SoK: Certified robustness for deep neural networks}.
\newblock {\em Proceedings - IEEE Symposium on Security and Privacy}, 2023-May:1289--1310, 9 2023.

\bibitem{Liang2018}
Bin Liang, Hongcheng Li, Miaoqiang Su, Pan Bian, Xirong Li, and Wenchang Shi.
\newblock Deep text classification can be fooled.
\newblock In {\em IJCAI International Joint Conference on Artificial Intelligence}, 2018.

\bibitem{Liao2022AreNetworks}
Hsuan-Cheng Liao, Chih-Hong Cheng, Maximilian Kneissl, and Alois Knoll.
\newblock Are attention networks more robust? towards exact robustness verification for attention networks.
\newblock In {\em Computer Safety, Reliability, and Security: 41st International Conference}, 2 2022.

\bibitem{Liu2016_star}
Wei Liu, Chaofeng Chen, Kwan-Yee~K Wong, Zhizhong Su, and Junyu Han.
\newblock {STAR-Net: A spatial attention residue network for scene text recognition}.
\newblock {\em The British Machine Vision Conference}, 2016.

\bibitem{Madry2018}
Aleksander Madry, Aleksandar Makelov, Ludwig Schmidt, Dimitris Tsipras, and Adrian Vladu.
\newblock Towards deep learning models resistant to adversarial attacks.
\newblock {\em 6th International Conference on Learning Representations}, 6 2018.

\bibitem{Mishra2012}
Anand Mishra, Karteek Alahari, and C~V Jawahar.
\newblock Scene text recognition using higher order language priors.
\newblock {\em The British Machine Vision Association}, 2012.

\bibitem{Mo2022}
Yichuan Mo, Dongxian Wu, Yifei Wang, Yiwen Guo, and Yisen Wang.
\newblock When adversarial training meets vision transformers: Recipes from training to architecture.
\newblock {\em Advances in Neural Information Processing Systems}, 35, 10 2022.

\bibitem{muller2021scaling}
Christoph Müller, François Serre, Gagandeep Singh, Markus Püschel, and Martin Vechev.
\newblock Scaling polyhedral neural network verification on gpus.
\newblock {\em Proceedings of the 4 th MLSys Conference}, 7 2021.

\bibitem{Muller2021}
Mark~Niklas Müller, Gleb Makarchuk, Gagandeep Singh, Markus Püschel, and Martin Vechev.
\newblock {PRIMA: General and precise neural network certification via scalable convex hull approximations}.
\newblock {\em Proceedings of the ACM on Programming Languages}, 6, 3 2021.

\bibitem{Olivier2021}
Raphael Olivier and Bhiksha Raj.
\newblock Sequential randomized smoothing for adversarially robust speech recognition.
\newblock {\em EMNLP 2021 - 2021 Conference on Empirical Methods in Natural Language Processing, Proceedings}, pages 6372--6386, 11 2021.

\bibitem{Paszke2019PyTorch:Library}
Adam Paszke, Sam Gross, Francisco Massa, Adam Lerer, James Bradbury, Gregory Chanan, Trevor Killeen, Zeming Lin, Natalia Gimelshein, Luca Antiga, Alban Desmaison, Andreas K{\"{o}}pf, Edward Yang, Zach DeVito, Martin Raison, Alykhan Tejani, Sasank Chilamkurthy, Benoit Steiner, Lu~Fang, Junjie Bai, and Soumith Chintala.
\newblock {PyTorch: An imperative style, high-performance deep learning library}.
\newblock {\em Advances in Neural Information Processing Systems}, 32, 12 2019.

\bibitem{Phan2013}
Trung~Quy Phan, Palaiahnakote Shivakumara, Shangxuan Tian, and Chew~Lim Tan.
\newblock Recognizing text with perspective distortion in natural scenes.
\newblock {\em Proceedings of the IEEE International Conference on Computer Vision}, pages 569--576, 2013.

\bibitem{Raghunathan2018}
Aditi Raghunathan, Jacob Steinhardt, and Percy Liang.
\newblock Semidefinite relaxations for certifying robustness to adversarial examples.
\newblock {\em Advances in Neural Information Processing Systems}, 2018-December:10877--10887, 11 2018.

\bibitem{Risnumawan2014}
Anhar Risnumawan, Palaiahankote Shivakumara, Chee~Seng Chan, and Chew~Lim Tan.
\newblock A robust arbitrary text detection system for natural scene images.
\newblock {\em Expert Systems with Applications}, 41:8027--8048, 12 2014.

\bibitem{Ryou2020ScalableNetworks}
Wonryong Ryou, Jiayu Chen, Mislav Balunovic, Gagandeep Singh, Andrei Dan, and Martin Vechev.
\newblock Scalable polyhedral verification of recurrent neural networks.
\newblock {\em Computer Aided Verification}, 12759 LNCS:225--248, 2020.

\bibitem{salman2019convex}
Hadi Salman, Greg Yang, Huan Zhang, Cho-Jui Hsieh, and Pengchuan Zhang.
\newblock A convex relaxation barrier to tight robustness verification of neural networks.
\newblock {\em Advances in Neural Information Processing Systems}, 32, 2019.

\bibitem{Scaman2018}
Kevin Scaman and Aladin Virmaux.
\newblock {Lipschitz regularity of deep neural networks: Analysis and efficient estimation}.
\newblock In {\em Advances in Neural Information Processing Systems}, volume 2018-Decem, pages 3835--3844. Neural information processing systems foundation, 2018.

\bibitem{Shi2017}
Baoguang Shi, Xiang Bai, and Cong Yao.
\newblock An end-to-end trainable neural network for image-based sequence recognition and its application to scene text recognition.
\newblock {\em IEEE Transactions on Pattern Analysis and Machine Intelligence}, 39(11):2298--2304, 2017.

\bibitem{Shi2016}
Baoguang Shi, Xinggang Wang, Pengyuan Lyu, Cong Yao, and Xiang Bai.
\newblock Robust scene text recognition with automatic rectification.
\newblock {\em Proceedings of the IEEE Computer Society Conference on Computer Vision and Pattern Recognition}, 2016-December:4168--4176, 3 2016.

\bibitem{Shi2023}
Zhouxing Shi, Qirui Jin, Huan Zhang, Zico Kolter, Suman Jana, and Cho-Jui Hsieh.
\newblock Formal verification for neural networks with general nonlinearities via branch-and-bound.
\newblock {\em The second Workshop on Formal Verification of Machine Learning, ICML}, 2023.

\bibitem{Shi2020RobustnessTransformers}
Zhouxing Shi, Huan Zhang, Kai-Wei Chang, Minlie Huang, and Cho-Jui Hsieh.
\newblock {Robustness verification for transformers}.
\newblock {\em Proceedings of the International Conference on Learning Representations}, 2 2020.

\bibitem{singh2019abstract}
Gagandeep Singh, Timon Gehr, Markus P{\"u}schel, and Martin Vechev.
\newblock An abstract domain for certifying neural networks.
\newblock {\em Proceedings of the ACM on Programming Languages}, 3(POPL):1--30, 2019.

\bibitem{Salzer2021}
Marco Sälzer and Martin Lange.
\newblock Reachability is np-complete even for the simplest neural networks.
\newblock {\em International Conference on Reachability Problems}, 13035 LNCS:149--164, 8 2021.

\bibitem{Tjeng2017EvaluatingProgramming}
Vincent Tjeng, Kai Xiao, and Russ Tedrake.
\newblock {Evaluating Robustness of Neural Networks with Mixed Integer Programming}.
\newblock {\em 7th International Conference on Learning Representations, ICLR 2019}, 11 2017.

\bibitem{vaswani2017attention}
Ashish Vaswani, Noam Shazeer, Niki Parmar, Jakob Uszkoreit, Llion Jones, Aidan~N Gomez, {\L}ukasz Kaiser, and Illia Polosukhin.
\newblock Attention is all you need.
\newblock {\em Advances in neural information processing systems}, 30, 2017.

\bibitem{Wang2011}
Kai Wang, Boris Babenko, and Serge Belongie.
\newblock End-to-end scene text recognition.
\newblock {\em Proceedings of the IEEE International Conference on Computer Vision}, pages 1457--1464, 2011.

\bibitem{wang2018efficient}
Shiqi Wang, Kexin Pei, Justin Whitehouse, Junfeng Yang, and Suman Jana.
\newblock Efficient formal safety analysis of neural networks.
\newblock {\em Advances in neural information processing systems}, 31, 2018.

\bibitem{Wang2021Beta-CROWN:Verification}
Shiqi Wang, Huan Zhang, Kaidi Xu, Xue Lin, Suman Jana, Cho-Jui Hsieh, and J.~Zico Kolter.
\newblock {Beta-CROWN: Efficient bound propagation with per-neuron split constraints for complete and incomplete neural network nobustness verification}.
\newblock {\em Conference on Neural Information Processing Systems}, 3 2021.

\bibitem{Wei2023ConvexVerification}
Dennis Wei, Haoze Wu, Min Wu, Pin-Yu Chen, Clark Barrett, and Eitan Farchi.
\newblock {Convex bounds on the Softmax function with applications to robustness verification}.
\newblock {\em International Conference on Artificial Intelligence and Statistics}, 2023.

\bibitem{weng2018towards}
Lily Weng, Huan Zhang, Hongge Chen, Zhao Song, Cho-Jui Hsieh, Luca Daniel, Duane Boning, and Inderjit Dhillon.
\newblock Towards fast computation of certified robustness for relu networks.
\newblock In {\em International Conference on Machine Learning}, pages 5276--5285. PMLR, 2018.

\bibitem{Wojna2017Attention-BasedImagery}
Zbigniew Wojna, Alexander~N. Gorban, Dar~Shyang Lee, Kevin Murphy, Qian Yu, Yeqing Li, and Julian Ibarz.
\newblock Attention-based extraction of structured information from street view imagery.
\newblock In {\em Proceedings of the International Conference on Document Analysis and Recognition, ICDAR}, 2017.

\bibitem{wong2018provable}
Eric Wong and Zico Kolter.
\newblock Provable defenses against adversarial examples via the convex outer adversarial polytope.
\newblock In {\em International conference on machine learning}, pages 5286--5295. PMLR, 2018.

\bibitem{Wu2019}
Min Wu and Marta Kwiatkowska.
\newblock {Robustness guarantees for deep neural networks on videos}.
\newblock {\em Proceedings of the IEEE}, 2020.

\bibitem{Xu2020AutomaticBeyond}
Kaidi Xu, Zhouxing Shi, Huan Zhang, Yihan Wang, Kai~Wei Chang, Minlie Huang, Bhavya Kailkhura, Xue Lin, and Cho~Jui Hsieh.
\newblock Automatic perturbation analysis for scalable certified robustness and beyond.
\newblock {\em Advances in Neural Information Processing Systems}, 2020-December, 2 2020.

\bibitem{Xu2021FastVerifiers}
Kaidi Xu, Huan Zhang, Shiqi Wang, Yihan Wang, Suman Jana, Xue Lin, and Cho~Jui Hsieh.
\newblock Fast and complete: Enabling complete neural network verification with rapid and massively parallel incomplete verifiers.
\newblock {\em 9th International Conference on Learning Representations}, 11 2021.

\bibitem{Yuan2020AdaptiveRecognition}
Xiaoyong Yuan, Pan He, Xiaolin Lit, and Dapeng Wu.
\newblock Adaptive adversarial attack on scene text recognition.
\newblock {\em Conference on Computer Communications Workshops}, pages 358--363, 7 2020.

\bibitem{Zhang2018}
H.~Zhang, T.-W. Weng, P.-Y. Chen, C.-J. Hsieh, and L.~Daniel.
\newblock {Efficient neural network robustness certification with general activation functions}.
\newblock In {\em Advances in Neural Information Processing Systems}, volume 2018-, pages 4939--4948. Neural information processing systems foundation, 2018.

\bibitem{Zhang2019RecurJac:Applications}
Huan Zhang, Pengchuan Zhang, and Cho-Jui Hsieh.
\newblock Recurjac: An efficient recursive algorithm for bounding jacobian matrix of neural networks and its applications.
\newblock {\em Proceedings of the Thirty-Third AAAI Conference}, 2019.

\bibitem{Zhang2020}
Wei~Emma Zhang, Quan~Z. Sheng, Ahoud Alhazmi, and Chenliang Li.
\newblock Adversarial attacks on deep-learning models in natural language processing.
\newblock {\em ACM Transactions on Intelligent Systems and Technology}, 11(3):1--41, 2020.

\end{thebibliography}

\clearpage
\appendix

\part{Appendix} 


\localtableofcontents


\section{Background on Polyhedral Verification Bounds}\label{appendix:bounds}

In this section, for the sake of completeness we provide polyhedral bounds for the relevant layers in OCR/STR networks. 
While the bounds for these layers, apart from the STN bilinear map, are mostly covered in the literature, we include them as background for the modifications and extensions we applied. Note that,  since fully-connected (linear) layers can be transformed exactly, and CNN layers can be transformed into fully-connected layers~\cite{singh2019abstract}, we only cover polyhedral bounds for non-linear layers.

\subsection{STN Bilinear Map}\label{appendix:bilinear}

Recall the bilinear map, where $f(\cdot)$ is the ReLU function:
\begin{align}
I^r_{ci}=\sum_{n,m}^{H,W}I_{cnm}f(1-|p_{ix}-m|)f(1-|p_{iy}-n|).
\end{align}

Here, we provide detailed polyhedral bounds for $r_{ix}\coloneqq f(1-|p_{ix}-m|)$ from interval bounds $l_{ix}<p_{ix}<u_{ix}$. Since the function mapping $p_{ix}$ to $r_{ix}$ consists of four pieces of linear functions, and $l_{ix},u_{ix}$ can be in either of the four pieces, we enumerate all possible cases here.
\begin{itemize}
\item If $l_{ix},u_{ix}$ are within the same linear piece, let $\nabla$ be the gradient of the piece of linear function $l_{ix}$ and $u_{ix}$ are both inside ($\nabla=0,1 or -1$), then $\nabla\leq r_{ix}\leq \nabla$.
\item If $[l_{ix},u_{ix}]$ are across exactly two linear pieces, then:
\begin{enumerate}
\item If $l_{ix}\in(-\infty,-1)$ and $u_{ix}\in [-1,0]$, then $u_{ix}(p_{ix}+1- l_{ix})/(u_{ix} - l_{ix})\leq r_{ix}\leq a_1 p_{ix}+a_1$ where $a_{1}\in[0,1]$.
\item If $l_{ix}\in[0,1)$ and $u_{ix}\in [1,\infty)$, then $u_{ix}(1-p_{ix}- l_{ix})/(u_{ix} - l_{ix})\leq r_{ix}\leq a_2 p_{ix}-a_2$ where $a_{2}\in[-1,0]$.
\item If $l_{ix}\in[-1,0)$ and $u_{ix}\in[0,1]$, then
\begin{equation}
\frac{-(u_{ix}+l_{ix})}{u_{ix}-l_{ix}}p_{ix}+\frac{2u_{ix}l_{ix}}{u_{ix}-l_{ix}}+1\leq r_{ix}\leq a_{3}p_{ix}+1, \text{ where } a_{3}\in[-1,1].
\end{equation}
\end{enumerate}
\item If $[l_{ix},u_{ix}]$ are across exactly three linear pieces, then:
\begin{enumerate}
    \item If $l_{ix}\in[-1,0]$ and $u_{ix}\in[1,\infty)$, then
    \begin{equation}
    \frac{-(1+l_{ix})}{1-l_{ix}}p_{ix}+\frac{2l_{ix}}{1-l_{ix}}+1\leq r_{ix}\leq a_{4}p_{ix}+1, \text{ where } a_{4}\in[-1/u_{ix},1].
    \end{equation}
    \item If $l_{ix}\in(-\infty,-1]$ and $u_{ix}\in[0,1]$, then
    \begin{equation}
    \frac{-(u_{ix}-1)}{u_{ix}+1}p_{ix}+\frac{-2u_{ix}}{u_{ix}+1}+1\leq r_{ix}\leq a_{5}p_{ix}+1, \text{ where }a_{5}\in[-1,-1/l_{ix}].
    \end{equation}
\end{enumerate}
\item Finally, if $[l_{ix},u_{ix}]$ are across all four linear pieces, then $0\leq r_{ix}\leq a^u_{3}p_{ix}+1$, where $a^u_{3}\in[-1/u_{ix},-1/l_{ix}]$.
\end{itemize}

This concludes all possible cases for $l_{ix}$ and $u_{ix}$. Bounds for $r_{iy}\coloneqq f(1-|p_{iy}-n|)$ can be derived similarly, and with the polyhedral bounds of $I_{cnm}$, $r_{ix}$ and $r_{iy}$, the final polyhedral bounds for $\image^r$ can be computed via addition and multiplication bounds (\ref{appendix:multi}).

\subsection{ReLU}\label{appendix:relu}

\noindent To the best of our knowledge, polyhedral bounds for the ReLU activation function $\sigma(x) = \max\{x, 0\}$ were first introduced in \cite{singh2019abstract}. Given an input $x \in [l_x, u_x]$ and with polyhedral upper and lower bounds $a^\geq(x), a^\leq(x)$, they distinguish the following three cases:
\begin{enumerate}
    \item If $u_x \leq 0$, then $a^\leq(\sigma(x)) = a^\geq(\sigma(x)) = 0$ and $l_{\sigma(x)} = u_{\sigma(x)} = 0$.
    \item If $0 \leq l_x$, then $a^\geq(\sigma(x)) = a^\leq(\sigma(x)) = x$, $l_{\sigma(x)} = l_x$ and $u_{\sigma(x)} = u_x$.
    \item Otherwise, they set $a^\geq(\sigma(x)) = u_x (x - l_x)/(u_x - l_x)$ and $a^\leq(\sigma(x)) = \lambda x$ for $\lambda \in [0, 1]$. In practice, they choose the $\lambda$ that minimizes the area between the upper and lower bounds in the $(x, \sigma(x))$-plane. Finally, they let $l_{\sigma(x)} = \lambda l_x$ and $u_{\sigma(x)} = u_x$.
\end{enumerate}

\subsection{Multiplication}\label{appendix:multi}

\noindent To bound the product of two scalar variables $x \in [l_x, u_x]$ and $y \in [l_y, u_y]$ under perturbation, \cite{Shi2020RobustnessTransformers} use lower and upper polyhedral planes parameterized by coefficients $A_l, B_l, C_l$ and $A_u, B_u, C_u$, i.e.
\begin{equation}
    A_lx + B_ly + C_l \leq xy \leq A_ux + B_uy + C_u 
\end{equation}

\noindent They show that choosing $A_l = l_y, A_u = u_y, B_l = B_u = l_x, C_l = -l_xl_y$, and $C_u = -l_xu_y$ is optimal in that this choice of parameters minimizes the integrals of $F^L(x, y)$ and $F^U(x, y)$ over $[l_x, u_x] \times [l_y, u_y]$, where
\begin{equation}
\begin{split}
    F^L(x, y) &= xy - \left(A_lx + B_ly + C_l\right) \\
    F^U(x, y) &= xy - \left(A_ux + B_uy + C_u\right)
\end{split}
\end{equation}

\noindent For the proof of this result, see \cite{Shi2020RobustnessTransformers}. \\

\subsection{LSTM}



We adopt Prover~\cite{Ryou2020ScalableNetworks} for certifying LSTM layers, and we briefly introduce their polyhedral bounds in this section. The idea behind the LSTM architecture is to handle long-term sequential dependencies, for example in words or sentences. These dependencies are passed through time with two vectors, a cell state $\cell^{(t)}$ and a hidden state $\hidden^{(t)}$ for every timestep $t$. The state vectors are updated using the following equations:
\begin{equation}
\begin{split}
&f_0^{(t)} = [x^{(t)},\hidden^{(t-1)}]\weight_f+\bias_f\\
&o_0^{(t)} = [x^{(t)},\hidden^{(t-1)}]\weight_o+\bias_o\\
&c^{(t)} = \sigma(f_0^{(t)})\odot c^{(t-1)}+\sigma(i_0^{(t)})\odot\tanh(\tilde{c}_0^{(t)})\\
  \end{split}
\quad\quad\quad
  \begin{split}
&i_0^{(t)} = [x^{(t)},\hidden^{(t-1)}]\weight_i+\bias_i\\
&\tilde{c}_0^{(t)} = [x^{(t)},\hidden^{(t-1)}]\weight_{\tilde{c}}+\bias_{\tilde{c}}\\
&h^{(t)}=\sigma(o_0^{(t)})\odot\tanh(c^{(t)})
  \end{split}
\end{equation}

where $[\cdot, \cdot]$ denotes the horizontal concatenation of two row vectors, $W$ and $b$ denote the kernel and bias of the cell, and $\sigma$ denotes the sigmoid function. \\

\textbf{Polyhedral Verification of LSTM:}
\cite{Ryou2020ScalableNetworks} bound the products of the identity, sigmoid and tanh functions using lower and upper polyhedral planes parameterized by coefficients $A_l, B_l, C_l$ and $A_u, B_u, C_u$. For 
$$h(x,y)=\begin{cases}\sigma(x)\tanh(y)\\
\sigma(x)y
\end{cases},$$ it follows that
\begin{equation}
    A_l\cdot x+B_l\cdot y+C_l\leq h(x,y)\leq A_u\cdot x+B_u\cdot y+C_u
\end{equation}

\noindent The problem of finding the bounding polyhedral planes can be reduced to an optimization problem. For the lower bound
\begin{align}
    \min_{A_l,B_l,C_l}\int_{(x,y)\in B} (h(x,y)-(A_l\cdot x+B_l\cdot y+C_l))\nonumber\\
    \text{subject to }A_l\cdot x+B_l\cdot y+C_l\leq h(x,y), \quad\forall(x,y)\in B \label{op_objective}
\end{align}
where $B=[l_x,u_x]\times[l_y,u_y]$ is the input boundary region of neurons $x$ and $y$. To solve this optimization problem, \cite{Ryou2020ScalableNetworks} first \textbf{approximate this intractable optimization problem using Monte Carlo sampling via LP}. Let $D={(x_1,y_1),...,(x_n,y_n)}$ be the uniformly sampled points at random from $B$, the approximation of the objective in Equation~ \ref{op_objective} is
\begin{align}
\min_{A_l,B_l,C_l}\sum_{i=1}^n (h(x_i,y_i)-(A_l\cdot x_i+B_l\cdot y_i+C_l))\nonumber\\
    \text{subject to }\bigwedge_{i=1}^n A_l\cdot x_i+B_l\cdot y_i+C_l\leq h(x_i,y_i).\label{approx_objective}
\end{align}

\noindent This provides potentially unsound bounds, meaning that there can be points in region $B$ that violate
the bounds. To provide soundness, one can \textbf{adjust the offset to guarantee soundness utilizing Fermat’s
theorem}. Next, compute $\Delta_l=\min_{(x,y)\in B} h(x,y)-(A_l\cdot x+B_l\cdot y+C_l)$ and adjust the lower bound by updating the offset: $C_l\gets C_l+\Delta_l$. To compute $\Delta_l$, let $A_l\cdot x+B_l\cdot y+C_l$ be the initial lower bound in $B$ obtained from the LP approximation. 

\noindent
For $h(x,y)=\sigma(x)\tanh(y)$, consider the extreme points of $F(x,y)=\sigma(x)\tanh(y)-(A_l\cdot x+B_l\cdot y+C_l)$ via its partial derivatives:
\begin{align}
    &\frac{\partial F}{\partial x}=\sigma(x)\tanh(y)(1-\sigma(x))-A_l\\
    &\frac{\partial F}{\partial y}=\sigma(x)(1-\tanh^2(y))-B_l
\end{align}

\noindent Consider three different cases: 

\noindent1. If $x\in\{l_x,u_x\}$ and $y\in[l_y,u_y]$, $\frac{\partial F}{\partial y}=0$ can be written as 
\begin{equation}
    (1-\tanh^2(y))=B_l/S_x\label{f:case1}
\end{equation}
where $S_x=\sigma(x)$ is a constant.\\

\noindent2. If $x\in[l_x,u_x]$ and $y\in\{l_y,u_y\}$, setting $\frac{\partial F}{\partial x}=0$ becomes
\begin{equation}
    \sigma(x)(1-\sigma(x)=A_l/T_y\label{f:case2}
\end{equation}
where $T_y=\tanh(y)$ is a constant.\\

\noindent3. Otherwise, consider both $\frac{\partial F}{\partial x}=0$ and $\frac{\partial F}{\partial y}=0$ to reduce $\tanh(y)$ and obtain 
\begin{equation}
    \sigma(x)^4+(-2-B_l)\sigma(x)^3+(1+2B_l)\sigma(x)^2+(-B_l)\sigma(x)-A_l^2=0.\label{f:case3}
\end{equation}

\noindent According to Fermat's theorem on stationary points, $F(x,y)$ achieves its extremum at $B$ either in the roots of Equation~ \ref{f:case1},  \ref{f:case2} and  \ref{f:case3} or at the four corners of $B$. Thus, one can obtain $\Delta_l=\min_{(x,y)\in B} F(x,y)$ by evaluating $F$ at these points and selecting the minimum among them.\\

\noindent For $h(x,y)=\sigma(x)y$, the analysis is similar and can be found in \cite{Ryou2020ScalableNetworks}.








\begin{subsection}{Normalization Layer}

\noindent Normalize each vector $v_j$ by subtracting the mean of its entries, i.e. 
$$v_{j + 1}^i = v_j^i - \text{mean}(v_j) = v_j - \frac{1}{N_j} \sum_{i = 1}^{N_j} v_j^i$$

\noindent where $N_j$ is the dimension of $v_j$ for some $j > 1$, and $v_j^i$ is its $i$-th entry. Note that the normalization layer is a linear layer, in that the normalized vector $v_{j + 1}$ can be expressed as 
$$v_{j + 1} = v_j - \frac{1}{N_j}\textbf{1}v_j = \left(I - \frac{1}{N_j}\textbf{1} \right) v_j$$

\noindent where \textbf{1} denotes the $N_j \times N_j$ matrix of all ones and $I$ is the $N_j \times N_j$ identity matrix. Hence the associated abstract transformer is exact, i.e. the polyhedral upper and lower bounds agree and are given by the above expression.\\ 


\end{subsection}




\begin{subsection}{Tanh}

Following \cite{singh2019abstract}, let $l_j$ and $u_j$ be the concrete lower (resp. upper) bounds in the previous layer. Then we set $l_i' = \tanh(l_j)$ and $u_i' = \tanh(u_j)$. If $l_j = u_j$, then $a_i'^{\leq}(x) = a_i'^{\geq}(x) = \tanh(l_j)$. Otherwise, consider $a_i'^{\leq}(x)$ and $a_i'^{\geq}(x)$ separately. Let 
$$ \lambda = \frac{\tanh(u_j) - \tanh(l_j)}{u_j - l_j}$$

\noindent and
$$ \lambda' = \min \left\{ \tanh'(l_j), \tanh'(u_j)\right\}$$

\noindent If $0 < l_j$, then $a_i'^{\leq}(x) = \tanh(l_j) + \lambda (x_j - l_j)$, otherwise $a_i'^{\leq}(x) = \tanh(l_j) + \lambda' (x_j - l_j)$. If $u_j \leq 0$, then $a_i'^{\geq}(x) = \tanh (u_j) + \lambda (x_j - u_j)$, and $a_i'^{\geq}(x) = \tanh (u_j) + \lambda' (x_j - u_j)$ otherwise.\\

\subsection{Polyhedral Verification for the dot product and Softmax}

We considered three possible ways to bound the softmax function presented in the literature \cite{Shi2020RobustnessTransformers, Ryou2020ScalableNetworks, Wei2023ConvexVerification}, of which we adopt the third approach as its bounds are provably tighter than the bounds derived from the other two methods \cite{Wei2023ConvexVerification}.

\subsubsection{Bounding multiplication, division, and exponential separately}

\cite{Shi2020RobustnessTransformers} derive polyhedral bounds for multiplication, division, and for the exponential function. The desired bounds follow using function composition and linear transformations described below. \\

\begin{itemize}
    \item \textbf{Multiplication:} Let $x$ and $y$ be scalars with upper and lower bounds $u_x, u_y$ and $l_x, l_y$. Then
    $$ l_y x + l_x y - l_x l_y \leq x y \leq u_y x + l_x y - l_x u_y$$
    \item \textbf{Division:} Using the same setting as above, decompose the division operation as $x/y = x \cdot 1/y$, so that it suffices to derive bounds for the reciprocal function. Assume that $0 < l_y \leq y \leq u_y$ and denote $1/y =: \sigma(y)$. Then
    $$ \sigma'\left(\frac{u_y + l_y}{2} \right) \left[ y - \left( \frac{u_y + l_y}{2} \right)\right] + \sigma \left( \frac{u_y + l_y}{2}\right) \leq \sigma(y) \leq \frac{\sigma(u_y) - \sigma(l_y)}{u_y - l_y} \left( y - l_y\right) + \sigma(l_y)$$
    \item \textbf{Exponential:} Proceeding as above,
    $$ \exp(d) \left(x - d \right) + \exp(d) \leq \exp(x) \leq \frac{\exp(u_x) - \exp(l_x)}{u_x - l_x} \left( x - l_x\right) + \exp(l_x)$$

    \noindent In the lower bound, let $d := \min \left( (l_x + u_x)/2, l + 1 - \Delta_d \right)$, where $\Delta_d$ is a small positive number, such as $10^{-2}$.\\
\end{itemize}

\subsubsection{Polyhedral Verification for Division using Fermat} We can improve the bounds for division using Fermat's theorem as in \cite{Ryou2020ScalableNetworks}. For the lower bound, we assume that $l_x \leq x \leq u_x$ and $0 < l_x \leq y \leq u_y$ and define
$$ F^L(x, y) = \frac{x}{y} - \left(\alpha^L  x + \beta^L y + \gamma^L \right)$$

\noindent where $\alpha, \beta$ and $\gamma$ are real numbers. The partial derivatives are given by
$$ \frac{\delta F^L}{\delta x} = \frac{1}{y} - \alpha^L \text{, } \frac{\delta F^L}{\delta y} = - \frac{x}{y^2} - \beta^L$$

\noindent We distinguish between the following cases:
\begin{enumerate}
    \item If $x \in \{l_x, u_x\}$, then $\frac{\delta F^L}{\delta y} = 0$ implies that $y^2 = -\frac{x}{\beta^L}$ (recall that $y > 0$ by assumption as in the previous subsection).\\
    \item If $y \in \{l_y, u_y\}$, then $\frac{\delta F^L}{\delta y}$ is constant, so $F$ is monotonous on the two boundaries, hence it suffices to consider the corner points.\\
\end{enumerate}

\noindent We also note that there are no (isolated) minima inside the boundaries, i.e. within $(l_x, u_x) \times (l_y, u_y)$, because,  for that to be the case, $F^L$ would have to be positive definite at some point. However,
$$ \frac{\delta^2 F^L}{\delta x^2} = 0 \text{, } \frac{\delta^2 F^L}{\delta y^2} = \frac{2x}{y^3} \text{, } \frac{\delta^2 F^L}{\delta x \delta y} = -\frac{1}{y^2}$$

\noindent Hence for the determinant of the Hessian of $F^L$, we have
$$ \frac{\delta^2 F^L}{\delta x^2} \cdot \frac{\delta^2 F^L}{\delta y^2} - \left( \frac{\delta^2 F^L}{\delta x \delta y}\right) = -\frac{1}{y^4} < 0$$

\noindent so that the eigenvalues of the Hessian cannot have the same sign, hence $F^L$ cannot be positive definite.\\

\noindent In conclusion, to obtain the offset needed for a sound lower bound, it suffices to ensure that $F^L \geq 0$ at the corners of $[l_x, u_x] \times [l_y, u_y]$ and at the root (w.r.t. $y$ and with $x = \{l_x, u_x\}$) of $y^2 = -\frac{x}{\beta^L}$, if it exists and lies within the region. \\

\noindent \textbf{Computing the Lower Bound:} Proceeding by analogy with \cite{Ryou2020ScalableNetworks}, we first find an approximate a lower bound by solving
$$ \text{min}_{\alpha^L, \beta^L, \gamma^L} \sum_{i = 1}^n \left[ \frac{x_i}{y_i} - \left(\alpha^L x_i + \beta^L y_i + \gamma^L \right)\right]$$

\noindent where the $x_i$ and $y_i$ are sampled from $B = [l_x, u_x] \times [l_y, u_y]$. We use the parameters $\alpha^L, \beta^L$, and $\gamma^L$ found in this way to define the function $F^L(x, y)$ above, and let $\Delta^L = \min_{(x, y) \in B} F^L(x, y)$, for which it suffices to consider the four corners and the points $(l_x, \sqrt{-l_x / \beta^L}), (u_x, \sqrt{-u_x / \beta^L})$, if they exist (in $B$). The resulting sound bound is then
$$ \alpha^L x + \beta^L y + \gamma^L + \Delta^L$$

\end{subsection}

\subsubsection{Improved Softmax Bounds}

Finally, the provably tighter bounds for softmax are introduced in \cite{Wei2023ConvexVerification}. They define
$$p_j = \frac{1}{1 + \sum_{j' \neq j} \exp\left(x_{j'} - x_j\right)}$$

\noindent for $x \in \mathbb{R}^K$ and $j = 1, ..., K$. For ease of notation, they focus on $j = 1$. All other cases follow by symmetry. They derive the following lower and upper bounds on the softmax output $p_1$:
\begin{equation*}
\begin{split}
    L^{\text{LSE}}(x) &= \frac{\exp(x_1)}{\overline{\text{SE}}(x; l, u)}, \\
    U^{\text{LSE}}(x) &= \frac{\underline{p}_1 \log(\overline{p}_1) - \overline{p}_1 \log(\underline{p}_1) - (\overline{p}_1 - \underline{p}_1) \text{LSE}(\tilde{x})}{\log(\overline{p}_1) - \log(\underline{p}_1)},
\end{split}
\end{equation*}

\noindent where
\begin{equation*}
\begin{split}
    \overline{\text{SE}}(x; l, u) &= \sum_{j = 1}^K \left( \frac{u_j - x_j}{u_j - l_j}\exp{l_j} + \frac{x_j - l_j}{u_j - l_j}\exp{u_j}\right), \\
    \text{LSE}(\tilde{x}) &= \log \left( \sum_{j = 1}^K \exp{x_j}\right), \\
    \overline{p}_1 &= \frac{1}{1 + \sum_{j \neq 1} \exp(\tilde{l}_j)}, \\
    \underline{p}_1 &= \frac{1}{1 + \sum_{j \neq 1} \exp(\tilde{u}_j)}.
\end{split}
\end{equation*}

\noindent Here, $\tilde{x}_j = x_j - x_1$, $\tilde{u}_j = u_j - l_1$, and $\tilde{l}_j = l_j - u_1$. They linearize $L^{\text{LSE}}(x)$ and $U^{\text{LSE}}(x)$ by using tangent planes to these bounds. Recall that, for a function $f: \mathbb{R}^K \rightarrow \mathbb{R}$, a tangent plane at a point $c \in \mathbb{R}^K$ can be described as
$$ \overline{f}_c(x) = \sum_{j = 1}^K \left( \frac{\delta f(c)}{\delta x_j} (x_j - c_j)\right) + f(c)$$

\noindent or in the linearized form we use in our implementation,
$$ \overline{f}_c(x) = \sum_{j = 1}^K \frac{\delta f(c)}{\delta x_j} x_j - \sum_{j = 1}^K \frac{\delta f(c)}{\delta x_j} c_j + f(c)$$

\noindent In our case,
\begin{equation*}
\begin{split}
    \frac{\delta L^{\text{LSE}}(x)}{\delta x_1} &= L^{\text{LSE}}(x) - \exp(x_1) \frac{1}{\overline{\text{SE}}(x; l, u)^2} \left( \frac{\exp(u_1) - \exp(l_1)}{u_1 - l_1}\right), \\
    \frac{\delta L^{\text{LSE}}(x)}{\delta x_i} &= - \exp(x_1) \frac{1}{\overline{\text{SE}}(x; l, u)^2} \left( \frac{\exp(u_i) - \exp(l_i)}{u_i - l_i}\right), i \neq 1, \\
    \frac{\delta U^{\text{LSE}}(x)}{\delta x_1} &= -\frac{\overline{p}_1 - \underline{p}_1}{\log(\overline{p_1}) - \log(\underline{p_1})} \left(\frac{\exp(x_1)}{\text{SE}(x)} - 1 \right),\\
    \frac{\delta U^{\text{LSE}}(x)}{\delta x_i} &= -\frac{\overline{p}_1 - \underline{p}_1}{\log(\overline{p_1}) - \log(\underline{p_1})} \frac{\exp(x_i)}{\text{SE}(x)}, i \neq 1,
\end{split}
\end{equation*}

\noindent where $\text{SE}(x) = \sum_{j = 1}^K \exp(x_j)$.

\section{Models and Training}\label{appendix:arch}

In this section, we introduce the detailed model architectures, training process, datasets, and hardware specifications.

\subsection{Model Architectures}
We did not use lexicon for all models, and we convert all images to gray-scale single-channel images scaled to $20 \times 100$. For the standard STR model pipelines, the image is input into the TPS transformation module described in \ref{tab:TPS_arch}, where the linear projection and the grid generator creates the grid for the transformation, and the grid sampler is the bilinear sampler that produces the rectified image. The rectified image is then put into the feature extractor, sequence modelling and decoder head depicted in \ref{tab:standard_acrh}, where \ref{tab:ctc_acrh} shows the detailed architecture of the CTC decoder model and \ref{tab:attn_acrh} shows the detailed architecture of the attention decoder model. For the ViTSTR model, the architecture is given in \ref{tab:vitstr_arch}. The input image is directly input into the model, where the patch embedding cuts the image into $5\times5$ pieces and is linearly projected to 128 hidden dimensions, before adding the positional encoding. We use 5 layers of the Transformer block for the results in the main paper, followed by reshaping and linear projection to output the sequence of labels. For all models, the number of output frames varies from 11 to 21 depending on the model, whereas and the number of classes is 42, which includes the alphabet, numbers and some punctuation symbols.

\begin{table}[t]
    \centering
    \tiny
    \subfloat[CTC decoder model architecture]{
    \begin{tabular}{||c|c||}
    \hline
    \textbf{Layer Type} & \textbf{Configuration}  \\ [0.5ex]
    \hline \hline
    Input & $20\times100\times1$\\      \hline
    Conv + ReLU & $6\times6\times32$, s:2, p:0\\    \hline
    Conv + ReLU & $5\times5\times64$, s:1, p:2\\    \hline
    Max Pooling & $1\times2$, s:2\\    \hline
    Batch Norm & -\\    \hline
    Conv + ReLU & $3\times3\times128$, s:2, p:0\\    \hline
    Conv + ReLU & $3\times3\times128$, s:1, p:1\\    \hline
    Conv + ReLU & $3\times3\times128$, s:1, p:0\\    \hline
    Batch Norm & -\\    \hline
    \hline
    Reshape & $\#$frames$\times128$\\    \hline
    LSTM & $64$ hidden dim\\    \hline\hline
    Linear Projection & $\#$frames $\times\#$classes\\    \hline
    Softmax & -\\ \hline
    CTC Decoder & -\\\hline
    \end{tabular}\label{tab:ctc_acrh}}
    \hspace{10pt}
    \subfloat[Attention decoder model architecture]{
    \begin{tabular}{||c|c||}
    \hline
    \textbf{Layer Type} & \textbf{Configuration}  \\ [0.5ex]
    \hline \hline
    Input & $20\times100\times1$\\      \hline
    Conv + ReLU & $6\times6\times32$, s:2, p:0\\    \hline
    Conv + ReLU & $5\times5\times64$, s:1, p:2\\    \hline
    Max Pooling & $1\times2$, s:2\\    \hline
    Batch Norm & -\\    \hline
    Conv + ReLU & $3\times3\times128$, s:1, p:1\\    \hline
    Max Pooling & $1\times2$, s:2\\    \hline
    Conv + ReLU & $3\times3\times128$, s:1, p:1\\    \hline
    Conv + ReLU & $2\times2\times128$, s:1, p:0\\    \hline
    Batch Norm & -\\    \hline
    \hline
    Map to Sequence & $7\times128$\\    \hline
    LSTM & $64$ hidden dim\\    \hline\hline
    Linear Projections & 64 hidden dim\\ \hline
    Softmax & -\\ \hline
    Attention Mul &-\\ \hline
    LSTM & 64 hidden dim\\ \hline
    Generator & $\#$frames$\times\#$classes\\ \hline
    \end{tabular}\label{tab:attn_acrh}}
    \caption{The model architecture of the feature extractor, sequence modelling and decoder head for the standard STR models. Here \texttt{s} and \texttt{p} stand for stride and padding size, respectively.}
    \label{tab:standard_acrh}
\end{table}

\begin{table}[t]
    \centering
    \tiny
        \begin{tabular}{||c|c||}
    \hline
    \textbf{Layer Type} & \textbf{Configuration}  \\ [0.5ex]
    \hline \hline
    Input & $20\times100\times1$\\      \hline
    Conv + ReLU & $6\times6\times32$, s:2, p:0\\    \hline
    Max Pooling & $2\times2$, s:2\\    \hline
    Conv + ReLU & $5\times5\times64$, s:1, p:2\\    \hline
    Max Pooling & $2\times2$, s:2\\    \hline
    Conv + ReLU & $3\times3\times128$, s:1, p:1\\    \hline
    Average Pooling & 1 output dim\\ \hline
    Linear Projection & 128 hidden dim \\ \hline
    Grid Generator & -\\ \hline
    Grid Sampler & -\\ \hline


    \end{tabular}
    \caption{Model architecture for TPS transformation}
    \label{tab:TPS_arch}
\end{table}

\begin{table}[t]
    \centering
        \begin{tabular}{||c|c||}
    \hline
    \textbf{Layer Type} & \textbf{Configuration}  \\ [0.5ex]
    \hline \hline
    Input & $20\times100\times1$\\      \hline
    Patch Embedding & $5\times 5$\\    \hline
    Linear Projection & 128 hidden dim\\    \hline
    Positional Encoding & -\\    \hline\hline

    Transformer Blocks: & $\times$ 5\\      \hline
    Layer Norm & -\\    \hline
    Multi-Head Attention & 128 hidden dim\\    \hline
    Residule Connection & -\\    \hline
    Layer Norm & -\\    \hline
    Fully Connected + ReLU & 128 hidden dim\\    \hline
    Fully Connected + ReLU & 256 hidden dim\\    \hline
    Fully Connected + ReLU & 128 hidden dim\\    \hline
    Residule Connection & -\\    \hline\hline

    Reshape & $\#$frames$\times 128$\\    \hline
    Linear Projection & $\#$frames$\times\#$classes\\    \hline
    
    \end{tabular}
    \caption{Model architecture for the ViTSTR}
    \label{tab:vitstr_arch}
\end{table}

\subsection{Datasets}

{\bf MJSynth (MJ)}~\cite{Jaderberg2014} is a synthetic dataset designed for STR, containing 8.9M word images. The word generation process is as follows: font rendering; border and shadow rendering; background coloring; composition; applying projective distortions; blending with real-world images; and finally adding noise. The {\bf SynthText (ST)}~\cite{Gupta2016} dataset is another synthetically generated dataset and was originally designed for scene text detection. Nevertheless, it has also been used for STR by cropping word boxes from larger images. ST has 5.5M training data once the word boxes are cropped and filtered for non-alphanumeric characters. We use both MJ and ST for training (14.4M images in total).

For evaluation, we use 6 datasets. {\bf IIIT5K-Words}~\cite{Mishra2012} is the dataset crawled from Google image searches, which consists of 2,000 images for training and 3,000 images for evaluation. {\bf ICDAR2013 (IC13)}~\cite{Karatzas2013} was created for the ICDAR 2013 Robust Reading competition. We use images from the born-digital images task, which consists of 3564 images for training and 1439 images for evaluation. {\bf ICDAR2015 (IC15)}~\cite{Karatzas2015} was created for the ICDAR 2015 Robust Reading competition. We use images from the focused scene text task that are captured by Google Glasses while under the natural movement of the wearer. The benchmark contains 4,468 images for training and 2,077 images for evaluation. {\bf Street View Text (SVT)}~\cite{Wang2011} contains outdoor street images collected from Google Street View, which contains noisy, blurry, and/or low-resolution images. It consists of 257 images for training and 647 images for evaluation. {\bf SVT Perspective (SVTP)}~\cite{Phan2013} is also collected from Google Street View, in which many images contain perspective projections to mimic non-frontal viewpoints. It contains 645 images for evaluation. {\bf CUTE80 (CUTE)}~\cite{Risnumawan2014} is collected from natural scenes, of which many are curved text images. It contains 288 cropped images for evaluation. We use the training images from all these datasets as validation data for training, and we evaluated the models by applying STR-Cert on the evaluation images from these datasets.

\subsection{Training Configurations and Hyperparameters}
Experiments are carried out on a Linux server (Ubuntu 18.04.2) with two Intel Xeon Gold 6252 CPUs and six NVIDIA GeForce RTX 2080 Ti GPUs. All our algorithms are implemented in Python, where we adopt PyTorch~\cite{Paszke2019PyTorch:Library} for implementing the training and certification algorithms.

We mainly follow the training procedure of Baek \textit{et al.}~\cite{Baek2019}. We use the AdamW~\cite{Kingma2015} optimizer for training, with learning rate 0.001, betas=(0.9, 0.999) and weight decay=0.0001. The training batch is 512 and the total number of training iterations is 100k. PGD adversarial training~\cite{Madry2018} is adopted, where we perturb the image up to some \textit{perturbation budget} with 10 gradient ascent steps to maximize the prediction loss of the preturbed image. This adversarially attacked image is then fed into the neural network and trained using the standard training loss. Gradient clipping is used at magnitude 5. We validate the model every 1000 training iterations on the union of the training sets of IC13, IC15, IIIT, and SVT, to select the model with the highest accuracy on this set.

\section{Additional Experiments and Discussions}\label{appendix:exp}
In this section, we present additional experimental results for STR-Cert on STR models and provide a discussion. 

\subsection{Robustness Certification for Models with Different Depth}

We provide additional certification results for the CTC decoder models, attention decoder models, and ViTSTR with different depths compared to those used in the main paper. For the CTC and attention decoder models, we provide certification results for models with 6 convolutional layers in the feature extractor. For ViTSTR, we provide certification results for two additional models with different numbers of layers (4 and 6) of the Transformer block. In \ref{tab:exp_vitstr}, we present the percentage certified under various perturbation budgets. 
It can be seen that a higher number of layers reduces the percentage certified due to the increased depth for certification, where the error compounds. However, similar scalability trends with respect to the perturbation budget can be observed for all model depths. 

For the standard STR pipelines, we also experiment on models with different depths. The depth for the decoders is constant, and we found the depth of TPS transformation only marginally affects the accuracy and percentage certified. In addition, it is infeasible to certify models with more then 1 layer of LSTM, so we only study models with different numbers of convolutional layers in the feature extractor. In \ref{tab:exp_standard}, the results for CTC and attention decoder models with 6 layers of CNNs in the feature extractor are presented. A slight drop in percentage certified is observed, where the attention decoder model suffers more since the abstraction error from the feature extractor is compounded by certifying the two LSTMs in the model.

\begin{table}[t]
    \centering
    \tabcolsep=0.14cm
    \begin{tabular}{c|cccc|cccc}
    \toprule
    Model&\multicolumn{4}{c}{ViTSTR: 4 Transformer blocks}&\multicolumn{4}{c}{ViTSTR: 6 Transformer blocks} \\
    \cmidrule(lr){2-5}\cmidrule(lr){6-9}
    Datasets&$\epsilon=.001$ & $\epsilon=.003$ & $\epsilon=.005$ &$\epsilon=.01$ &$\epsilon=.001$ & $\epsilon=.003$ & $\epsilon=.005$ & $\epsilon=.01$\\
    \midrule
    IIIT5K & 97.5\% & 75.5\% & 57.0\% & 26.0\% & 92.5\% & 65.0\% & 45.0\% & 18.0\%\\
    IC13   & 98.5\% & 88.0\% & 74.5\% & 43.0\% & 95.5\% & 78.0\% & 59.5\% & 28.5\%\\
    IC15   & 95.5\% & 59.5\% & 41.5\% & 12.5\% & 91.5\% & 49.5\% & 34.5\% & 7.5\%\\
    SVT    & 97.5\% & 67.0\% & 60.5\% & 31.0\% & 92.5\% & 56.5\% & 45.5\% & 21.5\%\\
    SVTP   & 96.5\% & 62.0\% & 44.0\% & 22.0\% & 91.0\% & 50.0\% & 30.5\% & 13.5\%\\
    CUTE   & 97.0\% & 81.0\% & 59.5\% & 27.0\% & 92\% & 67.5\% & 45.0\% & 19.5\%\\
    \bottomrule
    \end{tabular}
        \caption{\% certified in the first 200 correctly classified instances for the ViTSTR model with different layers of Transformer blocks.}
    \label{tab:exp_vitstr}
\end{table}

\begin{table}[t]
    \centering
    \tabcolsep=0.14cm
    \begin{tabular}{c|ccc|ccc}
    \toprule
    Model&\multicolumn{3}{c}{CTC decoder: 6 layer CNNs} &\multicolumn{3}{c}{Attention decoder: 6 layer CNNs}\\
    \cmidrule(lr){2-4}\cmidrule(lr){5-7}
    Datasets&$\epsilon=.001$ & $\epsilon=.003$ & $\epsilon=.005$ &$\epsilon=.001$ & $\epsilon=.003$ & $\epsilon=.005$\\
    \midrule
    IIIT5K & 96.5\% & 73.5\% & 43.5\% & 87.0\% & 63.0\% & 33.0\% \\
    IC13   & 97.5\% & 86.0\% & 55.5\% & 91.5\% & 77.5\% & 48.5\% \\
    IC15   & 93.5\% & 53.5\% & 15.0\% & 87.5\% & 31.0\% & 8.5\% \\
    SVT    & 93.0\% & 65.0\% & 30.5\% & 84.0\% & 38.0\% & 16.5\% \\
    SVTP   & 92.5\% & 59.5\% & 31.5\% & 85.5\% & 53.5\% & 18.0\% \\
    CUTE   & 96.0\% & 81.5\% & 37.0\% & 90.5\% & 67.5\% & 28.5\% \\
    \bottomrule
    \end{tabular}
        \caption{\% certified in the first 200 correctly classified instances for CTC decoder and attention decoder model with 6 convolutional layers.}
    \label{tab:exp_standard}
\end{table}


\subsection{Robustness Against Rotation}

Adversarial robustness against rotations has been considered using DeepPoly~\cite{singh2019abstract}, where they certified inputs against the usual pointwise perturbation plus rotation angle $\theta$ within some range. The interval domain for each pixel under all rotation angles is used to create the adversarial region for the certification. This produces extremely wide pixel input intervals and effectively loses most of the information of the original image for large rotation intervals, making the final bounds too imprecise. To address this, Singh \textit{et al.}~\cite{singh2019abstract} proposed to refine the adversarial region by segmenting the rotation range into $n$ partitions and compute the adversarial region induced by each segment, before further segmenting each adversarial region's interval domain into $m$ parts and merging the certification results for all these $n*m$ adversarial regions. Although this method in theory allows for certification against rotation, it is very computationally intensive and sometimes impractical. Singh \textit{et al.}~\cite{singh2019abstract} used $n,m \approx 300$ to certify a single MNIST image against rotation between -45 to 65 degrees. For our case, we fail to certify a single sample against rotation within a reasonable computation time, and further increasing the granularity of the refinement would mean that each sample could take weeks to certify.

During experiments, we observed that, as expected, the STN module for the standard STR models is very robust against rotation, and the patch embedding for ViTSTR should also provide a certain level of robustness against rotation. Unfortunately, the rotation certification approach of DeepPoly is insufficient to prove that STR models are robust against rotation?. Therefore, developing certification tools specifically for rotation, perhaps based on an abstraction domain other than polyhedra, is a very interesting direction for future work.

\end{document}